\documentclass[lettersize,journal]{IEEEtran}
\usepackage{graphicx}
\usepackage{subfigure}
\usepackage{amsmath,amsfonts}
\usepackage{multirow}
\usepackage{comment} 
\usepackage{hhline}
\usepackage{tabularx}
\usepackage{booktabs}
\usepackage{multirow}
\usepackage{bigstrut}
\usepackage{amsmath} 
\usepackage{helvet}  
\usepackage{hyperref} 
\hypersetup{colorlinks=true,linkcolor=blue,anchorcolor=blue,citecolor=blue} 
\usepackage[table]{xcolor}
\usepackage{array}
\usepackage{algorithmic}
\usepackage{array}
\usepackage{textcomp}
\usepackage{cite}
\usepackage{stfloats}
\usepackage{url}
\usepackage{verbatim}
\usepackage{graphicx}
\usepackage{color}
\usepackage{bm}
\usepackage{mathrsfs}
\usepackage{amssymb}
\usepackage{float}
\usepackage{verbatim}
\usepackage{comment}
\def\BibTeX{{\rm B\kern-.05em{\sc i\kern-.025em b}\kern-.08em
    T\kern-.1667em\lower.7ex\hbox{E}\kern-.125emX}}
\usepackage{balance}

\begin{document}
\title{PC-NeRF: Parent-Child Neural Radiance Fields under Partial Sensor Data Loss in Autonomous Driving Environments}
\author{Xiuzhong Hu, Guangming Xiong, Zheng Zang, Peng Jia, Yuxuan Han*, Junyi Ma* %
\thanks{
	The research is funded by the National Natural Science Foundation of China under Grant 52372404. X. Hu, G. Xiong, Z. Zang, P. Jia, Y. Han, and J. Ma are with the School of Mechanical Engineering, Beijing Institute of Technology, Beijing, 100081, China. 
	
	*Corresponding author emails: yx\_han\_work@foxmail.com and junyi.ma@bit.edu.cn.}}

\maketitle

\begin{abstract}
Reconstructing large-scale 3D scenes is essential for autonomous vehicles, especially when partial sensor data is lost. Although the recently developed neural radiance fields (NeRF) have shown compelling results in implicit representations, the large-scale 3D scene reconstruction using partially lost LiDAR point cloud data still needs to be explored. To bridge this gap, we propose a novel 3D scene reconstruction framework called parent-child neural radiance field (PC-NeRF). The framework comprises two modules, the parent NeRF and the child NeRF, to simultaneously optimize scene-level, segment-level, and point-level scene representations. Sensor data can be utilized more efficiently by leveraging the segment-level representation capabilities of child NeRFs, and an approximate volumetric representation of the scene can be quickly obtained even with limited observations. With extensive experiments, our proposed PC-NeRF is proven to achieve high-precision 3D reconstruction in large-scale scenes. Moreover, PC-NeRF can effectively tackle situations where partial sensor data is lost and has high deployment efficiency with limited training time. Our approach implementation and the pre-trained models will be available at \url{https://github.com/biter0088/pc-nerf}.
\end{abstract}


\section{Introduction}
\begin{figure}[t]
	\centering
	\includegraphics[width=3.5in]{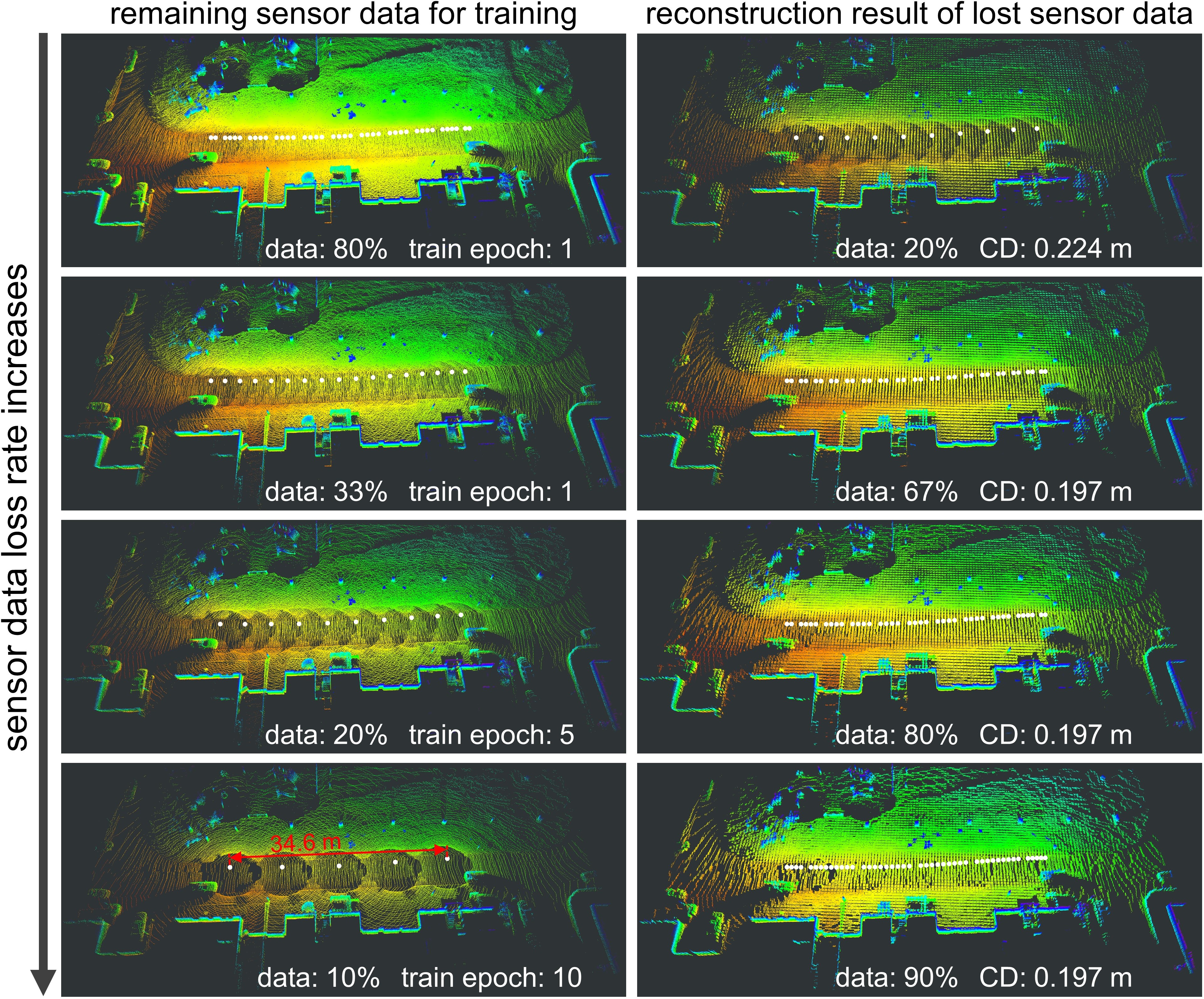}%
	\caption{With limited and even one-epoch training, our proposed PC-NeRF reconstructs 3D scenes well using LiDAR point cloud data from the KITTI 00 sequence at different sensor data loss rates. The white dots in each subfigure are the position of each scan, and the LiDAR operates at 10\,Hz. The sensor data loss rate of 20\% in the first row is a commonly used dataset division ratio, upon which we raise the challenge of 3D reconstruction by increasing the loss rate. Surprisingly, the 3D reconstruction accuracy of our proposed method increases when the sensor data loss rate increases from 20\% to 67\% because our devised model can rapidly learn an approximate scene distribution at the segment level. Moreover, if the input sensor data is too much or too little during training, more training epochs are needed to fit the model. More details are provided in Sec.~\ref{sec:Lesser point clouds for 3D reconstruction}.}	
	\label{fig:fig_motivation2}		
\end{figure}

\IEEEPARstart{R}{econstructing} a large-scale, high-precision 3D scene is essential for autonomous vehicles to conduct environmental exploration, motion planning, and closed-loop simulation \cite{li2023point, zhong2023shine, ran2023neurar, deng2023nerf, turki2022mega, chang2022lamp}. The environmental information capture and sensor data utilization are vital for 3D scene reconstruction. As existing autonomous systems have difficulty accomplishing complex tasks entirely autonomously, remote operators are often required to provide some assistance \cite{parr2023investigating, zhang2023hivegpt}. However, the sensor data transmitted back from the autonomous vehicles may be partially lost due to communication conditions that cannot be continuously ensured to be adequate \cite{wang2023data}. Moreover, the sensor equipment may fail to capture adequate environmental information due to hardware malfunction and adverse weather factors \cite{song2023synthetic, wang2021adaptive, raouf2022sensor, waqas2022automatic}. Therefore, besides environmental information capture, it is imperative to explore the sensor data utilization of the available limited sensor data for enhancing the perception capabilities of autonomous vehicles, as depicted in Fig. \ref{fig:fig_motivation2}.

By utilizing conventional explicit representations such as meshes, depth maps, and voxels, it becomes possible to depict the reconstructed scene visually \cite{hornung2013octomap, hu2023non, vizzo2021poisson}. However, as the explicit representation is discrete, it cannot represent the scene at unlimited resolution \cite{yang2023neural}. Besides, the explicit features are usually sparse, which further causes a performance drop when partial sensor data is lost in real applications. Unlike explicit representations, implicit representations typically use a spatial function to describe the scene geometry continuously \cite{yu2023sketch, wang2021neus} and can provide richer information about environmental features. As a trendy method of implicit representation, neural radiance fields (NeRF) \cite{mildenhall2021nerf} attract significant research interest in computer vision, robotics, and augmented reality communities \cite{liu2023efficient} since they provide a continuous representation of objects and environments with high resolution. NeRF represents a scene using a fully connected deep network, which maps a single continuous 5D coordinate (spatial location and viewing direction) to the volume density and view-dependent emitted radiance at that spatial location \cite{mildenhall2021nerf}.

Currently, most NeRF-related works have been carried out based on camera image data or indoor LiDAR scan data \cite{yang2023neural, sucar2021imap, moreau2022lens, zhu2022nice, turki2022mega, tancik2022block, zhenxing2022switch, rebain2021derf, reiser2021kilonerf, liu2020neural, wang2021neus, liu2023efficient, kuang2022ir}, with only a few works using LiDAR point cloud data in the outdoor environment \cite{deng2023nerf, tao2023lidarnerf, zhang2023nerf, Huang2023nfl}. In contrast to the camera, LiDAR is comparably robust to day-and-night light changes and different weather conditions \cite{ma2022overlaptransformer, wang2022performance}. Moreover, LiDAR can capture accurate distance information, which is extremely helpful for large-scale and high-accuracy outdoor mapping of the environment \cite{yu2023nf, deng2023nerf}. Therefore, it is significant to explore further reconstructing the outdoor environmental representation with LiDAR-based NeRF.

The main contributions of this paper are as follows:

$\bullet$ A large-scale and high-precision 3D scene reconstruction framework called parent-child neural radiance fields (PC-NeRF) effectively improves the reconstruction performance under partial sensor data loss in outdoor environments. We partition the entire autonomous vehicle driving environment into large blocks, denoted as parent NeRFs, and subdivide these blocks into geometric segments known as child NeRFs. This partition enables us to jointly optimize scene-level, segment-level, and point-level scene representations.

$\bullet$ To our knowledge, our proposed PC-NeRF is the first NeRF-based large-scale 3D scene reconstruction method using sparse and partially lost LiDAR point clouds, even though NeRF is a dense volumetric representation typically constructed using large amounts of sensor data. Our proposed PC-NeRF improves the data utilization efficiency of the remaining point cloud by constructing segment-level representations. 

$\bullet$ With extensive experiments, our proposed PC-NeRF has been validated for its high deployment efficiency. Training for just one epoch in most evaluation scenarios is sufficient to achieve high-precision 3D reconstruction results, even under severe partial sensor data loss conditions.

\section{Related Works}
With NeRF's inherent advantages of continuous dense volumetric representation, NeRF-based techniques in novel view synthesis \cite{tao2023lidarnerf, zhang2023nerf, Huang2023nfl, liu2020neural, rebain2021derf, reiser2021kilonerf}, scene reconstruction \cite{rematas2022urban, turki2022mega, tancik2022block, zhenxing2022switch, xiangli2022bungeenerf, liu2020neural, rebain2021derf, reiser2021kilonerf}, and localization systems \cite{sucar2021imap, moreau2022lens, zhu2022nice, deng2023nerf, kuang2022ir, wiesmann2023locndf} have rapidly developed and are highly referential and informative. NeRF's ability to model continuous scene representations is also an essential basis for our proposed PC-NeRF, which uses LiDAR data as inputs and divides the spaces into different scales. Therefore, this section reviews the literature on LiDAR-based NeRF and space-division-based NeRF.

\subsection{LiDAR-Based NeRF}
Motivated by NeRF's capability to render photo-realistic novel image views, several works have explored the potential of NeRF on LiDAR point cloud data for novel view synthesis \cite{tao2023lidarnerf, zhang2023nerf, Huang2023nfl} and robot navigation \cite{deng2023nerf, kuang2022ir, wiesmann2023locndf}.

The goal of novel view synthesis is to generate a view of a 3D scene from a viewpoint where no real sensor image has been captured, allowing the opportunity to observe real scenes from a virtual perspective. Neural Fields for LiDAR (NFL) \cite{Huang2023nfl} combines the rendering power of neural fields with a detailed, physically motivated model of the LiDAR sensing process, thus enabling it to accurately reproduce key sensor behaviors like beam divergence, secondary returns, and ray dropping. LiDAR-NeRF \cite{tao2023lidarnerf} converts the 3D point cloud into the range pseudo image in 2D coordinates and then optimizes three losses, including absolute geometric error, point distribution similarity, and realism of point attributes. Similar to LiDAR-NeRF, NeRF-LiDAR \cite{zhang2023nerf} has also employed the spherical projection strategy and consists of three key components: NeRF reconstruction of the driving scenes, realistic LiDAR point clouds generation, and point-wise semantic label generation. However, if more than one laser point is projected onto the same pseudo-pixel, only the one with the smallest distance is retained \cite{tao2023lidarnerf}. Besides, it is particularly noticeable when using small resolution range pseudo images, as spherical projections can lead to significant information loss \cite{milioto2019rangenet++}.  

In robotics, LiDAR-based NeRF is usually proposed for localization and mapping. IR-MCL \cite{kuang2022ir} focuses on the problem of estimating the robot's pose in an indoor environment using 2D LiDAR data. With the pre-trained network, IR-MCL can synthesize 2D LiDAR scans for an arbitrary robot pose through volume rendering. However, the error between the synthesized and real scans is relatively large. NeRF-LOAM \cite{deng2023nerf} presents a novel approach for simultaneous odometry and mapping using neural implicit representation with 3D LiDAR data. NeRF-LOAM employs sparse octree-based voxels combined with neural implicit embeddings, decoded into a continuous signed distance function (SDF) by a neural implicit decoder. However, NeRF-LOAM cannot currently operate in real-time with its unoptimized Python implementation. LocNDF \cite{wiesmann2023locndf} utilizes neural distance fields (NDFs) for robot localization, demonstrating the direct learning of NDFs from range sensor observations. LocNDF has raised the challenge of addressing real-time constraints, and our work endeavors to investigate this challenge.

In contrast to projecting the LiDAR point cloud onto a range pseudo-image, our proposed PC-NeRF handles 3D LiDAR point cloud data directly. Besides learning the LiDAR beam emitting process, our proposed PC-NeRF explores the deployment performance of NeRF-based methods.


\subsection{Space-Division-Based NeRF}
When large-scale scenes such as where the autonomous vehicles drive need to be represented with high precision, the model capacity of a single NeRF is limited in capturing local details with acceptable computational complexity \cite{turki2022mega, tancik2022block, zhenxing2022switch}. For large-scale 3D scene reconstruction tasks, Mega-NeRF \cite{turki2022mega}, Block-NeRF \cite{tancik2022block}, and Switch-NeRF \cite{zhenxing2022switch}  have adopted the NeRF sub-module solution. Mega-NeRF decomposes a scene into cells with centroids and initializes a corresponding set of model weights. At query time, Mega-NeRF produces opacity and color for a given position and direction using the model weights closest to the query point. Like Mega-NeRF, Block-NeRF \cite{tancik2022block} proposes dividing large environments into individually trained Block-NeRFs, which are then rendered and combined dynamically at inference time. For rendering a target view, a subset of the Block-NeRFs are rendered and then composited based on their geographic location compared to the camera. Switch-NeRF \cite{zhenxing2022switch} proposes a novel end-to-end large-scale NeRF with learning-based scene decomposition and designs a gating network to dispatch 3D points to different NeRF sub-networks. The gating network can be optimized with the NeRF sub-networks for different scene partitions by design with the Sparsely Gated Mixture of Experts.

Regarding space-division-based NeRF on a smaller scale, further space division is employed for faster and higher-quality rendering \cite{rebain2021derf, reiser2021kilonerf, liu2020neural}. DeRF \cite{rebain2021derf} proposes Voronoi Diagrams spatial decomposition and provides more efficient inference (with the same rendering quality) and higher-quality rendering (for the same inference cost) than the original NeRF \cite{mildenhall2021nerf}. KiloNeRF \cite{reiser2021kilonerf} demonstrates that real-time rendering is possible using thousands of tiny MLPs instead of one extensive multilayer perceptron network (MLP). Rather than representing the entire scene with a single, high-capacity MLP, KiloNeRF represents the scene with thousands of small MLPs. Neural Sparse Voxel Fields (NSVF) \cite{liu2020neural} define a set of voxel-bounded implicit fields organized in a sparse voxel octree to model local properties in each cell. Specifically, NSVF assigns a voxel embedding at each vertex of the voxel and obtains the representation of a query point inside the voxel by aggregating the voxel embeddings at the eight vertices of the corresponding voxel. The aggregating vertice voxel embedding is further passed through an MLP to predict the geometry and appearance of that query point.

Our work also draws on the ideas of spatial division on both large-scale and small-scale for autonomous driving scene representation. The proposed PC-NeRF divides the holistic driving spaces into multiple large blocks and further extracts the geometric segments in each block for detailed local representation.

\begin{figure*}[t]
	\centering
	\includegraphics[width=7.0in]{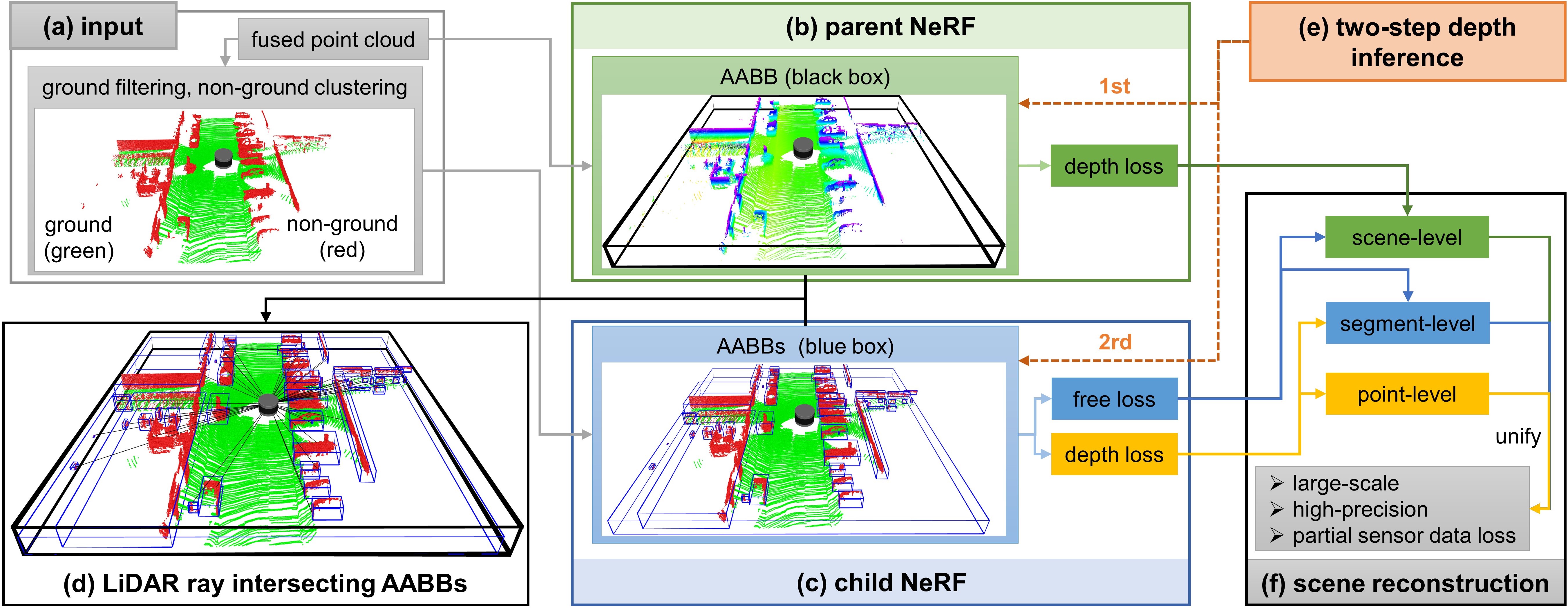}%
	\caption{Our PC-NeRF framework: (a) After multi-frame point cloud fusion, ground filtering, and non-ground point cloud clustering, we determine the spatial extent of the fused point cloud and geometric segments. (b) The parent NeRF depth loss supervises the volumetric representation of the parent NeRF Axis-Aligned Bounding Box (AABB). (c) The child NeRF depth loss supervises the space inside and around child NeRF AABBs, and the child NeRF free loss supervises the space around and far from these AABBs. (d) LiDAR rays intersect with parent NeRF and child NeRF AABBs. (e) For depth inference of each LiDAR ray, PC-NeRF searches in the parent NeRF AABB to locate corresponding child NeRF AABBs and then refines its inference in the child NeRF AABBs for higher precision. (f) Our PC-NeRF unifies scene-level, segment-level, and point-level information, enabling high-precision 3D reconstruction in large-scale scenes and addressing partial sensor data loss problems.}
	\label{framework}
\end{figure*}

\section{Parent-Child Neural Radiance Fields (PC-NeRF) Framework}
To further explore the large-scale high-precision 3D scene reconstruction based on LiDAR and NeRF and improve the reconstruction performance under partial sensor data loss, we propose a parent-child neural radiance field (PC-NeRF) framework, as shown in Fig. \ref{framework}. We divide the entire autonomous vehicle driving environment into large blocks, i.e., parent NeRFs, and geometric segments within the blocks, i.e., child NeRFs, to jointly optimize scene-level, segment-level, and point-level scene representation. With LiDAR point cloud data collected by autonomous vehicles as input, we design parent NeRF depth loss, child NeRF depth loss, and child NeRF free loss to train our PC-NeRF model. Moreover, we propose a two-step depth inference method to realize segment-to-point inference.

\subsection{Parent NeRF}\label{sec:parent NeRF}
To efficiently represent the autonomous vehicle's traversing environment, we construct multiple rectangular-shaped parent NeRFs (as Fig. \ref{framework}(b) illustrates) one after the other along the vehicle's trajectory, creating a new parent NeRF when the autonomous vehicle orientation variation exceeds a given threshold. Constructing parent NeRF requires three considerations: the effective LiDAR point cloud utilization, the driving environment representation, and the NeRF near/far bounds calculation. Since the point clouds become sparser the further away from the LiDAR origin while the NeRF model itself is a dense volumetric representation, laser points within a certain distance from the LiDAR origin rather than the holistic point clouds are chosen to train our proposed PC-NeRF model. Besides, the driving environments featured with roads, walls, and vehicles can be tightly enclosed by bounding boxes. Therefore, each parent NeRF's space is represented as a large Axis-Aligned Bounding Box (AABB) in our work, which also makes it easier to calculate the related near and far bounds for rendering. 

\subsection{Child NeRF}\label{sec:Child NeRFs}
To further represent the environment at the segment and point levels and efficiently capture the approximate environmental distributions under partial sensor data loss, we divide the point clouds in a parent NeRF into multiple child NeRFs that enclose each environmental geometric segment. Since the geometric segment quantity is limited and the child NeRFs' space can also be represented by AABBs, building child NeRFs from the raw point cloud is a fast way to generate detailed environmental representations. With less space volume, child NeRFs can represent a larger environment with the same model capacity by learning more detailed volumetric densities of segments. In addition, segment-level child NeRFs can compensate for the inadequate parts of the environmental representation caused by the sparsity of LiDAR point clouds.

The point cloud allocation for child NeRF can be divided into the following three steps, and the results are shown in Fig. \ref{framework}(a) and Fig. \ref{framework}(c). Step 1: Extract the ground point clouds from the fused point clouds related to road ground, roadside sidewalk, etc. The ground plane is one significant geometric segment in the driving environment and is spatially adjacent to the other individual geometric segments. Therefore, distinguishing it helps to extract other geometric segments further accurately. Step 2: Cluster the remaining non-ground point clouds into various segments using region-growing clustering. The advantages of the utilized clustering method are excellent scalability for operation on large-scale point clouds and the adaptive ability to cluster different shaped and sized objects. Step 3: Construct child NeRF AABBs. The AABBs of the segmented point clouds obtained from step 1 and step 2 can be used as child NeRFs' spatial extent. After the division process above, the fused point clouds are distributed into a limited quantity of child NeRFs.

\subsection{Training PC-NeRF}\label{sec:LiDAR loss}
In NeRF volume rendering, the depth value of a ray $\mathbf{r}$ is synthesized from the weighted sampling depth values $t$ between the near and far bounds $t_\mathrm{n}$ and $t_\mathrm{f}$ by:
\begin{equation}
	\label{equation2}
	d(\mathbf{r}) = \int_{t_\mathrm{n}}^{t_\mathrm{f}} w(t) \cdot t dt	
\end{equation}
where $\mathbf{r}(t) = \mathbf{o} + t\mathbf{d}$ represents a ray with camera or LiDAR origin $\mathbf{o}$ oriented as $\mathbf{d}$, and the volume rendering integration weights $w(t)$ is calculated by:
\begin{equation}
	\label{equation3}
	w(t) = \exp\left(-\int_{t_\mathrm{n}}^{t} \sigma(s)ds\right) \cdot \sigma(t)
\end{equation}
where $\exp(-\int_{t_n}^{t} \sigma(s)ds)$ is the visibility of $\mathbf{r}(t)$ from origin $\mathbf{o}$, and $\sigma(t)$ is the volumetric density at $\mathbf{r}(t)$.

\begin{figure}[H]
	\centering
	\includegraphics[width=3.5in]{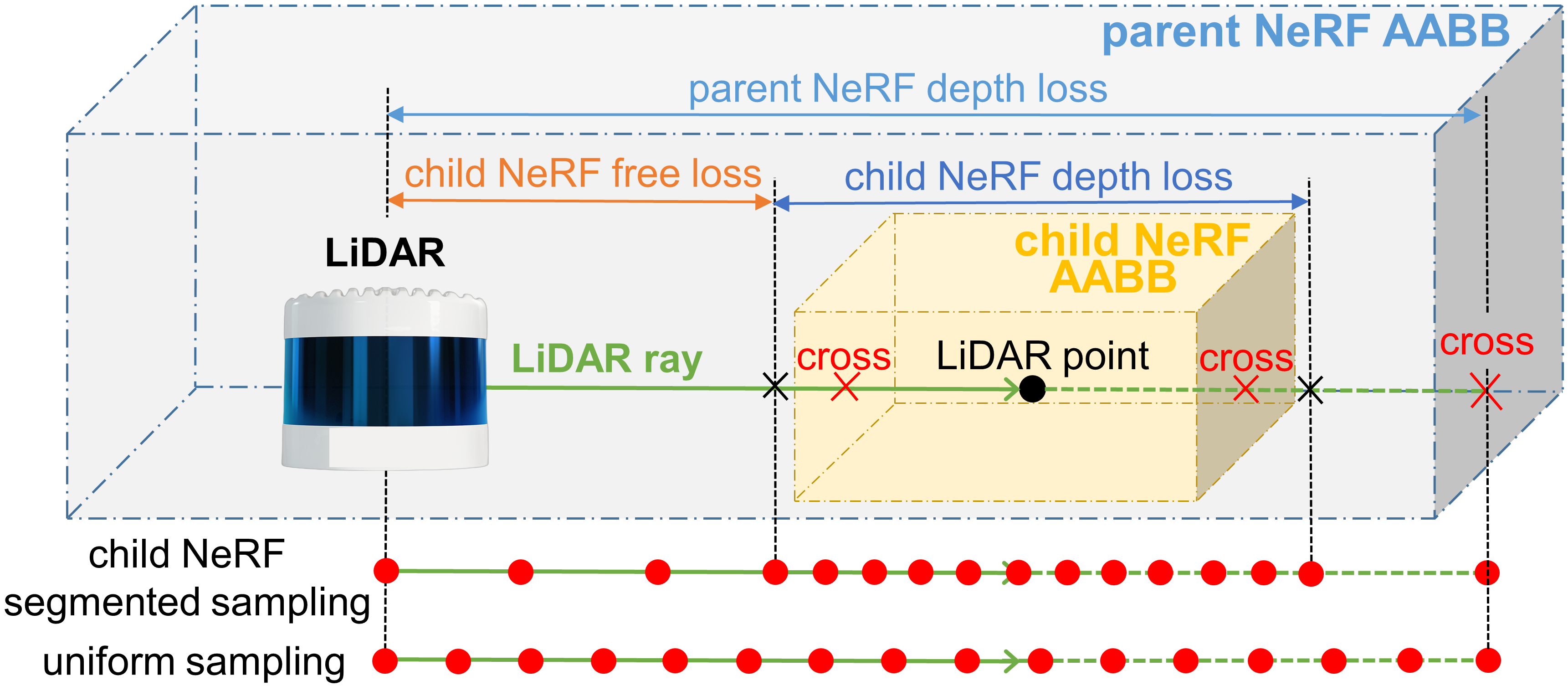}%
	\caption{Three LiDAR losses and child NeRF segmented sampling. The three LiDAR losses include parent NeRF depth loss, child NeRF depth loss, and child NeRF free loss. Using different sampling densities, the Child NeRF segmented sampling uniformly samples both inside and outside the intersection of the LiDAR ray with the Child NeRF.}
	\label{fig3}
\end{figure}

We apply LiDAR point cloud data to train our proposed PC-NeRF with model parameters $\bm{\theta}$. In the global coordinate system, the LiDAR origin position of the $i$-th frame is represented as $\mathbf{o}_{i} = (x_i, y_i, z_i)$. The depth, direction, and the corresponding child NeRF order number of the $j$-th laser point $\mathbf{p}_{ij} = (x_{ij}, y_{ij}, z_{ij})$ in the $i$-th point cloud are $d_{ij} = \|(\mathbf{p}_{ij}-\mathbf{o}_{i})\|_2$, $\mathbf{d}_{ij} = (\mathbf{p}_{ij}-\mathbf{o}_{i})/d_{ij}$ and $k_{ij}$ respectively. Then, we can get one LiDAR ray $\mathbf{r}_{ij} = [\mathbf{o}_{i}, \mathbf{p}_{ij}, d_{ij}, \mathbf{d}_{ij}, k_{ij}]$. We further design three LiDAR losses and child NeRF segmented sampling for our proposed PC-NeRF model, illustrated in Fig. \ref{fig3}.

\textbf{Parent NeRF depth loss}: In Sec.~\ref{sec:parent NeRF}, the autonomous vehicle driving environment is represented efficiently using parent NeRF. We use the intersection of the LiDAR ray $\mathbf{r}_{ij}$ with its corresponding parent NeRF surface as the far bound $f_{ij}^{p}$. To adequately represent the volumetric distribution of the whole driving environment, we set parent NeRF depth loss as:
\begin{equation}
	\label{equation7_1}
	\begin{aligned}
		\mathcal{L}_{ij}^{\mathrm{pd}}(\bm{\theta}) = 
		\mathcal{L}_{\mathrm{L1}}^{'}\left (\int_{t_0}^{f_{ij}^{p}} w(t) \cdot tdt, d_{ij}\right )		
	\end{aligned}
\end{equation}	
where the integration lower limit $t_0$ is set to 0. It is also recommended to set $t_0$ as $0.5\,m$ or $1\,m$ considering the space occupied by the LiDAR or the autonomous vehicle. And $\mathcal{L}_{\mathrm{L1}}^{'}(x, y) = 0.1\cdot \text{SmoothL1Loss}(10\cdot x, 10\cdot y)$ is an extension of $\text{SmoothL1Loss}$ that shifts the $|x-y|$ turning point from $1\,m$ to $0.1\,m$ to improve model sensitivity.

The commonly used depth inference approach uses the synthetic depth between the scene's near and far bounds as the inference depth \cite{mildenhall2021nerf, kuang2022ir, rematas2022urban, sucar2021imap, zhu2022nice}, which is used in Eq. \ref{equation7_1} and calculated as follows:
\begin{equation}
	\label{equation7_2}
	\begin{aligned}
		\hat{d_{ij}^1} = \int_{t_0}^{f_{ij}^{p}} w(t) \cdot tdt	
	\end{aligned}
\end{equation}	

\textbf{Segment-to-point hierarchical representation strategy}:
To find object surface points effectively and address the reduction of point cloud data, we further propose a segment-to-point hierarchical representation strategy. The depth difference $f_{ij}^{p} - t_0$ between parent NeRF near and far bounds in large-scale outdoor scenes can be about ten times greater than that in indoor scenes, making it more difficult for the NeRF model to find the object surface through sampling. Given that it is much easier to find segment-level child NeRF space than to find point-level object surface points between the parent NeRF near and far bounds, we propose a segment-to-point hierarchical representation strategy, which first finds the child NeRFs that intersect the LiDAR ray and then finds object surface points inside the child NeRFs that intersect the LiDAR ray.

We use the intersections of the LiDAR ray $\mathbf{r}_{ij}$ with its corresponding child NeRF surface as the child NeRF near and far bounds $[n_{ij}^{c}, f_{ij}^{c}]$. It is pretty evident that the depth difference between child NeRF near and far bounds $(f_{ij}^{c} - n_{ij}^{c})$ is much smaller than that between parent NeRF near and far bounds. Considering that the object surface has a certain thickness and the object surface may appear at the child NeRF bounds, we slightly inflate the child NeRF near and far bounds as $[n_{ij}^{c} - \varepsilon, f_{ij}^{c}+ \varepsilon]$, where $\varepsilon $ is a small inflation coefficient.

\textbf{Child NeRF free loss}: To make the process of finding the segment-level child NeRF's corresponding space faster, we propose the child NeRF free loss in Eq. \ref{equation5_1} as no opaque objects exist in $(t_0, n_{ij}^{c} - \varepsilon)$ and no objects can be observed in $(f_{ij}^{c}+ \varepsilon, f_{ij}^{p})$ in a large number of cases.
\begin{equation}
	\label{equation5_1}
	\begin{aligned}
		\mathcal{L}_{ij}^{\mathrm{cf}}(\bm{\theta}) = \int_{t_0}^{n_{ij}^{c} - \varepsilon}w(t)^2dt + \int_{f_{ij}^{c}+ \varepsilon}^{f_{ij}^{p}} w(t)^2dt
	\end{aligned}
\end{equation}

\textbf{Child NeRF depth loss}: Based on child NeRF free loss, we propose child NeRF depth loss to find object surface points inside the child NeRF, which is calculated by:

\begin{equation}
	\label{equation5}
	\begin{aligned}
		&\mathcal{L}_{ij}^{\mathrm{cd}}(\bm{\theta}) = 
		&\mathcal{L}_{\mathrm{L1}}^{'}\left (\int_{n_{ij}^{c} - \varepsilon}^{f_{ij}^{c}+ \varepsilon} w(t) \cdot tdt, d_{ij}\right )
	\end{aligned}
\end{equation}

To let the child NeRF free loss and the child NeRF depth loss have a smooth transition at the child NeRF bounds, $\mathcal{L}_{ij}^{\mathrm{cd}}(\bm{\theta})$ is further modified to:
\begin{equation}
	\label{equation5_2}
	\begin{aligned}
		&\mathcal{L}_{ij}^{\mathrm{cd}}(\bm{\theta}) = 
		&\mathcal{L}_{\mathrm{L1}}^{'}\left (\int_{n_{ij}^{c} - \varepsilon - \gamma}^{f_{ij}^{c}+ \varepsilon +\gamma} w(t) \cdot tdt, d_{ij}\right )
	\end{aligned}
\end{equation}
where $\gamma$ is a constant designed to represent the smooth transition interval on a LiDAR ray between the child NeRF free loss and the child NeRF depth loss.

To sum up, the total training loss from one LiDAR ray $\mathbf{r}_{ij}$ contains all the losses mentioned above:
\begin{equation}
	\label{equation8}
	\begin{aligned}
	\mathcal{L}_{ij}(\bm{\theta}) =  \lambda_{\mathrm{pd}}\mathcal{L}_{ij}^{\mathrm{pd}}(\bm{\theta}) + \lambda_{\mathrm{cf}} \mathcal{L}_{ij}^{\mathrm{cf}}(\bm{\theta}) + 
	\lambda_{\mathrm{cd}}\mathcal{L}_{ij}^{\mathrm{cd}}(\bm{\theta}) 
	\end{aligned}	
\end{equation}
where $\lambda_{\mathrm{pd}}$, $\lambda_{\mathrm{cf}}$ and $\lambda_{\mathrm{cd}}$ are the parameter to jointly optimize different losses.

\textbf{Child NeRF segmented sampling}: To efficiently find the objects in large-scale scenes even under partial sensor data loss, we propose a child NeRF segmented sampling method for objects more likely to be found in and around the child NeRFs. Assuming that $N$ points are sampled uniformly along the LiDAR rays, the child NeRF segmented sampling is sampling $\lambda_{\mathrm{in}} \cdot  N$ points in $[n_{ij}^{c} - \varepsilon, f_{ij}^{c} + \varepsilon] $ and sampling $( 1-\lambda_{\mathrm{in}}) \cdot  N$ in $[t_0, f_{ij}^{p}]$, as shown in Fig. \ref{fig3}. Therefore, child NeRF segmented sampling guarantees that at least $\lambda_{\mathrm{in}} \cdot  N$ points are sampled inside child NeRF, which means that at least $\lambda_{\mathrm{in}} \cdot  N$ sampling points are sampled near the real object.

\subsection{Two-step Depth Inference}\label{sec:Two-step Range Value Inference}
\begin{figure}[!t]
	\centering
	\subfigure[LiDAR rays intersect with different AABBs.]{\includegraphics[width=3.2in]{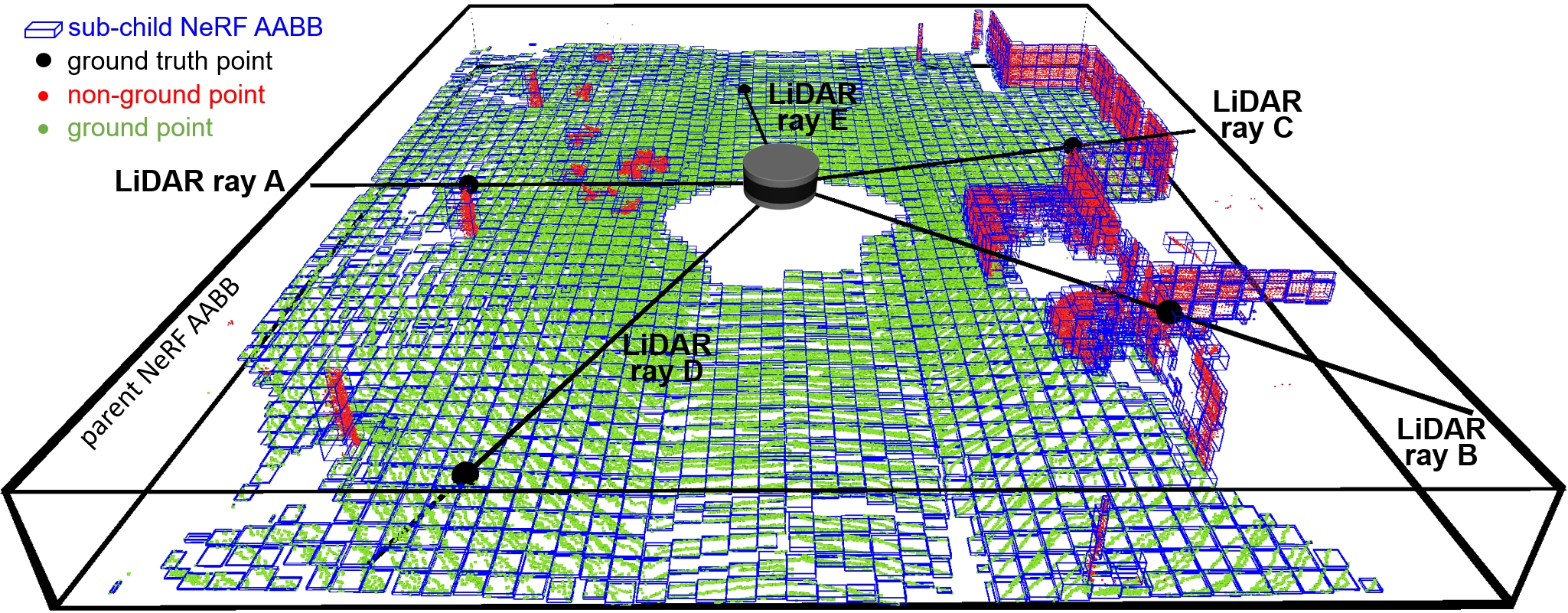}%
		\label{fig_infer_1}}
	\hspace{0.01in}		
	\subfigure[Weight distribution and depth inference along 5 LiDAR rays.]{\includegraphics[width=3.2in]{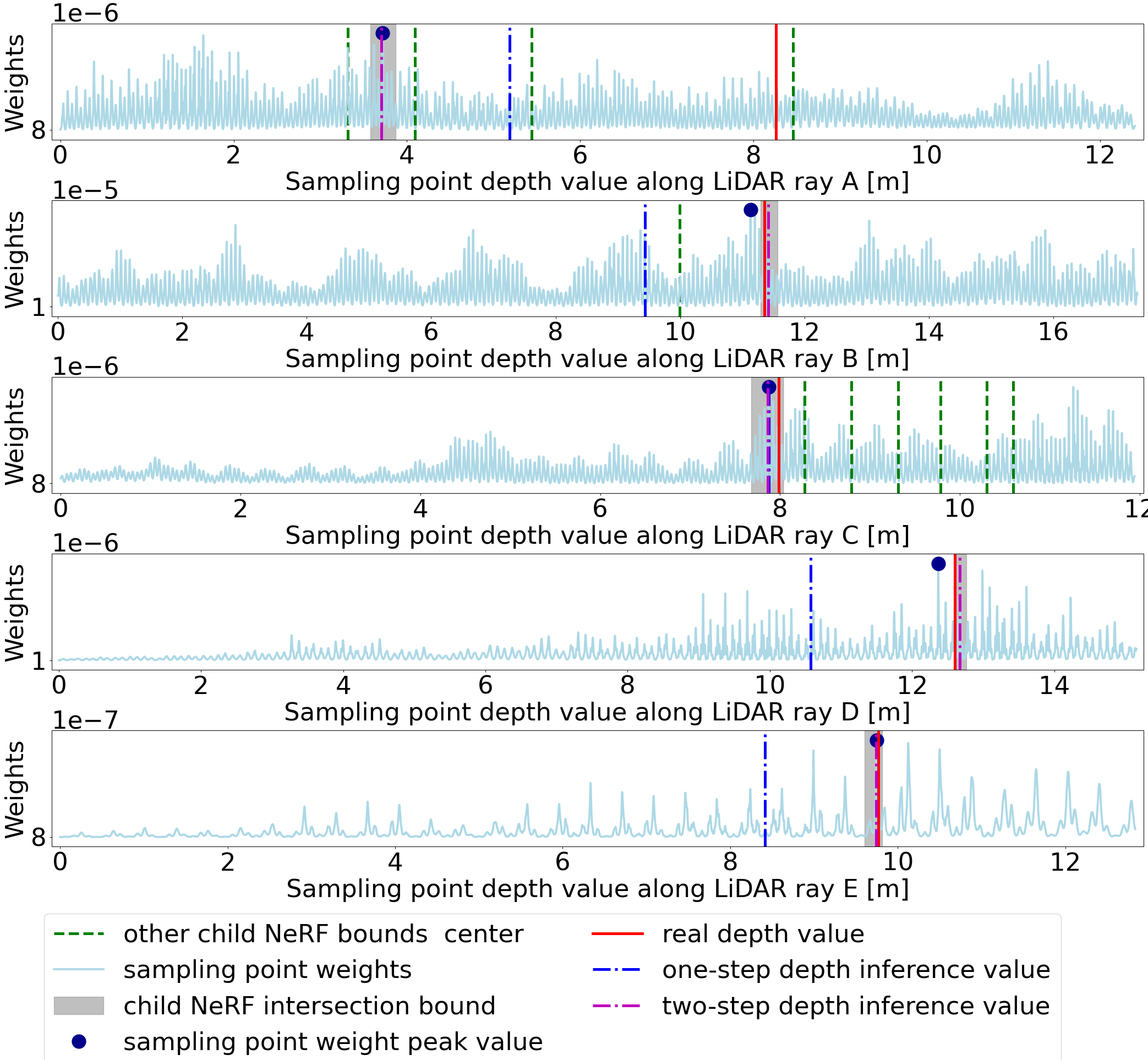}%
		\label{fig_infer_2}}
	\hfil	
	\caption{Parent-child NeRF's two-step depth inference effect illustration. The five subfigures in Fig. \ref{fig_infer}(b) represent depth value inference results for the five LiDAR rays in Fig. \ref{fig_infer}(a), where the weight distribution data comes from our proposed PC-NeRF model trained on the KITTI-sequence00-1151-1200 dataset in Sec.~\ref{sec:Evaluating}.}
	\label{fig_infer}
\end{figure}

Most current depth inference methods are one-step depth inference methods, similar to Eq. \ref{equation7_2}. In contrast, we provide a two-step depth inference method to infer more accurately based on the remaining sensor data. We first search in the parent NeRF AABB to acquire the child NeRF AABBs potentially intersecting the LiDAR ray $\mathbf{r}_{ij}$ and then conduct further inference in the child NeRF AABB's near and far bounds $[\hat{n_{ij}^{c}}, \hat{f_{ij}^{c}}]$ intersecting with the LiDAR ray, as shown in Fig. \ref{fig_infer}(a). To this end, we first select the child NeRF whose AABB outer sphere intersects the ray and then use the Axis Aligned Bounding Box intersection test proposed by Haines \cite{haines1989essential}, which can readily process millions of voxels or AABBs in real time \cite{liu2020neural}. If the LiDAR ray does not intersect any child NeRF AABB, the space extent of all child NeRF AABBs for this LiDAR ray should be slightly inflated in incremental steps, and the above inference should be performed again. 

To reduce the potentially severe impact of mistaking free space for object space, we keep the inferred depth values within one of the child NeRF near and far bounds, as shown in the first subfigure of Fig. \ref{fig_infer}(b). In real-world applications, the LiDAR ray usually intersects multiple child NeRF AABBs. In that case, we select the child NeRF AABB whose space area includes the weight (as Eq. \ref{equation3} calculates) peak value, as the LiDAR ray cannot penetrate opaque objects when it emits, as shown in the first and third subfigure of Fig. \ref{fig_infer}(b). Furthermore, suppose the peak weight value is not in any child NeRF AABB. In that case, the child NeRF with the maximum weight integration $W$, i.e., the most likely existence range of an object on the LiDAR ray, is selected, as shown in the second subfigure of Fig. \ref{fig_infer}(b). Moreover, the weight integration $W$ is calculated by Eq. \ref{equation9_1}. When the LiDAR ray only intersects one single child NeRF AABB, we can directly select it as the object area, as shown in the last two subfigures of Fig. \ref{fig_infer}(b). Note that the child NeRF weight integration $W$ may be minimal, in which case the LiDAR ray does not intersect any child NeRF and no depth value needs to be inferred. The inferred depth value of each LiDAR ray is calculated by Eq. \ref{equation10}.
\begin{equation}
	\label{equation9_1}
	W = \int_{\hat{n_{ij}^{c}}}^{\hat{f_{ij}^{c}}} w(t)dt
\end{equation}
\begin{equation}
	\label{equation10}
	\hat{d_{ij}} =\frac{{\int_{\hat{n_{ij}^{c}}}^{\hat{f_{ij}^{c}}}w(t)\cdot tdt}}
	{{\int_{\hat{n_{ij}^{c}}}^{\hat{f_{ij}^{c}}} w(t)dt}}
\end{equation}

\section{Experiments}
\subsection{Experiment Setups}
\textbf{Datasets}: We evaluate our approach using two publicly available outdoor datasets, including the MaiCity \cite{vizzo2021poisson} and KITTI odometry datasets \cite{geiger2012we}. The MaiCity dataset contains 64-beam noise-free synthetic LiDAR scans in virtual urban-like environments. KITTI odometry dataset also contains 64-beam LiDAR data collected by the vehicle in real-world urban environments and provides a localization benchmark. Moreover, we further use semantic labels from SemanticKITTI \cite{behley2019semantickitti} to filter out movable objects. In Sec.~\ref{sec:Evaluating} and Sec.~\ref{sec:Ablation}, we sample one test scan out of every five scans and use the other scans for training, the same as the dataset split in some previous works, such as NeRF-LOAM \cite{deng2023nerf} and NFL \cite{Huang2023nfl}.

\textbf{Metrics}: We evaluate the novel LiDAR view synthesis and the 3D Reconstruction performance of our method. For the novel LiDAR view synthesis, We compare the reconstructed depth value with the real depth of each LiDAR ray on the test set and report the mean average absolute error (Avg. Error [$\mathrm{m}$]) and mean accuracy at two thresholds (Acc@0.2$\mathrm{m}$ and Acc@1$\mathrm{m}$ [\%]). Moreover, for 3D reconstruction, we report the chamfer distance (CD [$\mathrm{m}$]) and F-score at two thresholds (F-score@0.2$\mathrm{m}$ and F-score@1$\mathrm{m}$). 

\textbf{Baselines}: A standard pipeline for generating new LiDAR views is constructing a 3D point cloud map and then using the ray-casting approach \cite{thrun2002probabilistic} to query new point clouds from the map. We implement this pipeline, which voxelizes the 3D point cloud map into a 3D voxel map to speed up the query but may slightly reduce accuracy, as a baseline method named \textbf{MapRayCasting}. In addition, we also extend the original NeRF model proposed by Mildenhall \cite{mildenhall2021nerf} by replacing a camera ray with a LiDAR ray as IR-MCL \cite{kuang2022ir}, named \textbf{OriginalNeRF}. 

\textbf{Training details}: We train our proposed PC-NeRF and all the baselines on an NVIDIA GeForce RTX 3090 and use Adam \cite{kingma2014adam} as the training optimizer. The initial learning rate is set to $4\times 10^{-5}$ and is adjusted using Pytorch's MultiStepLR strategy, where the adjustment milestones are $[5,10,20]$ and the adjustment factor is $0.1$. In Sec.~\ref{sec:Evaluating} and Sec.~\ref{sec:Lesser point clouds for 3D reconstruction}, $\lambda_{\mathrm{pd}}$, $\lambda_{\mathrm{cf}}$, $\lambda_{\mathrm{cd}}$, $\lambda_{\mathrm{in}}$, and $\gamma$ of our proposed PC-NeRF are set to $1$, $10^{6}$, $10^{5}$, $0.1$, and $2.0\,m$, respectively. This group of parameters is not specifically tuned for a single scene but is valid for all experimental scenes. Except as specifically noted, our proposed PC-NeRF uses only one training epoch to achieve the demonstrated accuracy. Both our PC-NeRF and OriginalNeRF use the hierarchical volume sampling strategy along the LiDAR ray proposed in NeRF \cite{mildenhall2021nerf}, where the points number $N_c$ and $N_f$ for coarse and fine sampling along the ray are 768 and 1536, respectively. However, for coarse sampling, our PC-NeRF uses the Child NeRF segmented sampling proposed in Sec.~\ref{sec:LiDAR loss}, while OriginalNeRF samples uniformly along the LiDAR rays.

\subsection{Evaluation for 3D Reconstruction in Different Scales}\label{sec:Evaluating}
\begin{figure*}[!t]
	\centering
	\fontfamily{phv}\selectfont 
	\resizebox{\textwidth}{!}{
	\noindent 
	\begin{tabularx}{\textwidth}{@{}c@{\hspace{0.005in}}c@{\hspace{0.005in}}c@{\hspace{0.005in}}c@{\hspace{0.05in}}c@{\hspace{0.05in}}c@{\hspace{0.05in}}c@{}}			
		\multicolumn{1}{c}{\fontsize{8}{8}\selectfont Scene}  & \multicolumn{1}{c}{\fontsize{8}{8}\selectfont Pointcloud Map}&
		\multicolumn{1}{c}{\fontsize{8}{8}\selectfont Scan}&		 \multicolumn{1}{c}{\fontsize{8}{8}\selectfont GT}& \multicolumn{1}{c}{\fontsize{8}{8}\selectfont MapRayCasting}& \multicolumn{1}{c}{\fontsize{8}{8}\selectfont OriginalNeRF}& \multicolumn{1}{c}{\fontsize{8}{8}\selectfont \textbf{PC-NeRF}} \\
		\multicolumn{1}{c}{\fontsize{8}{8}\selectfont \begin{tabular}[c]{@{}c@{}}MaiCity-\\sequence00\\-0-49\end{tabular}} 
		&\raisebox{-0.32\height}{\subfigure{\includegraphics[width=0.165\textwidth]{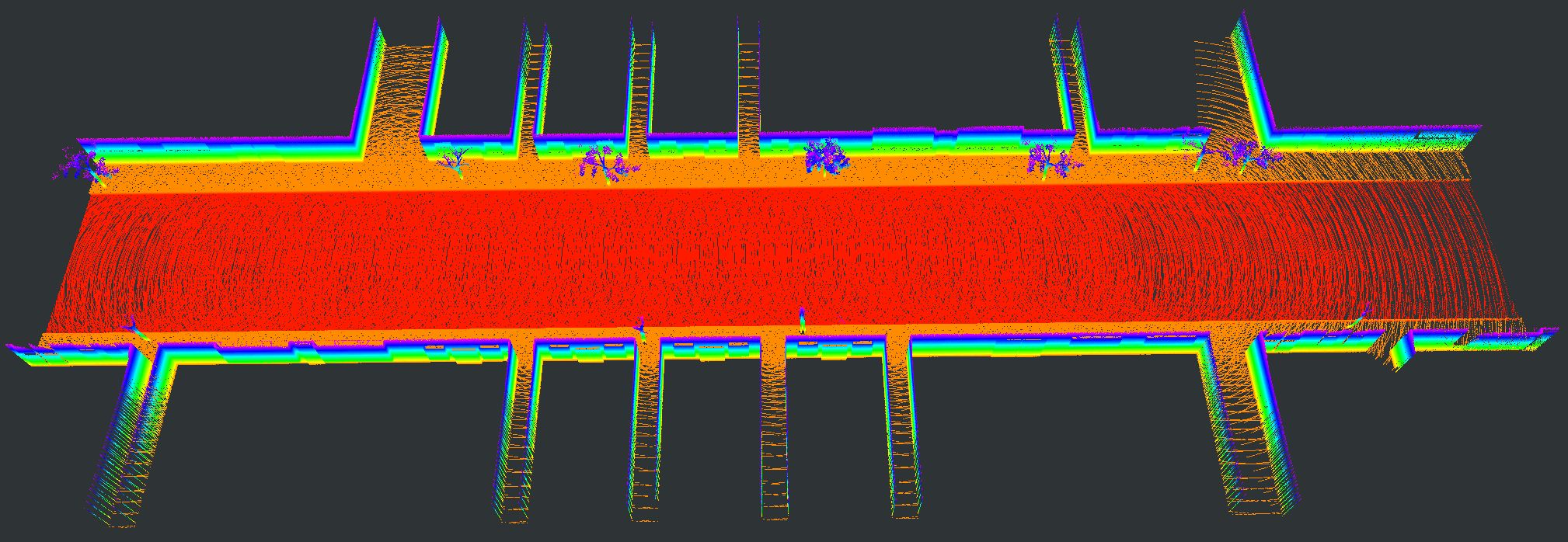}}} & 	\multicolumn{1}{c}{\fontsize{8}{8}\selectfont \begin{tabular}[c]{@{}c@{}}22\end{tabular}} & \raisebox{-0.32\height}{\subfigure{\includegraphics[width=0.165\textwidth]{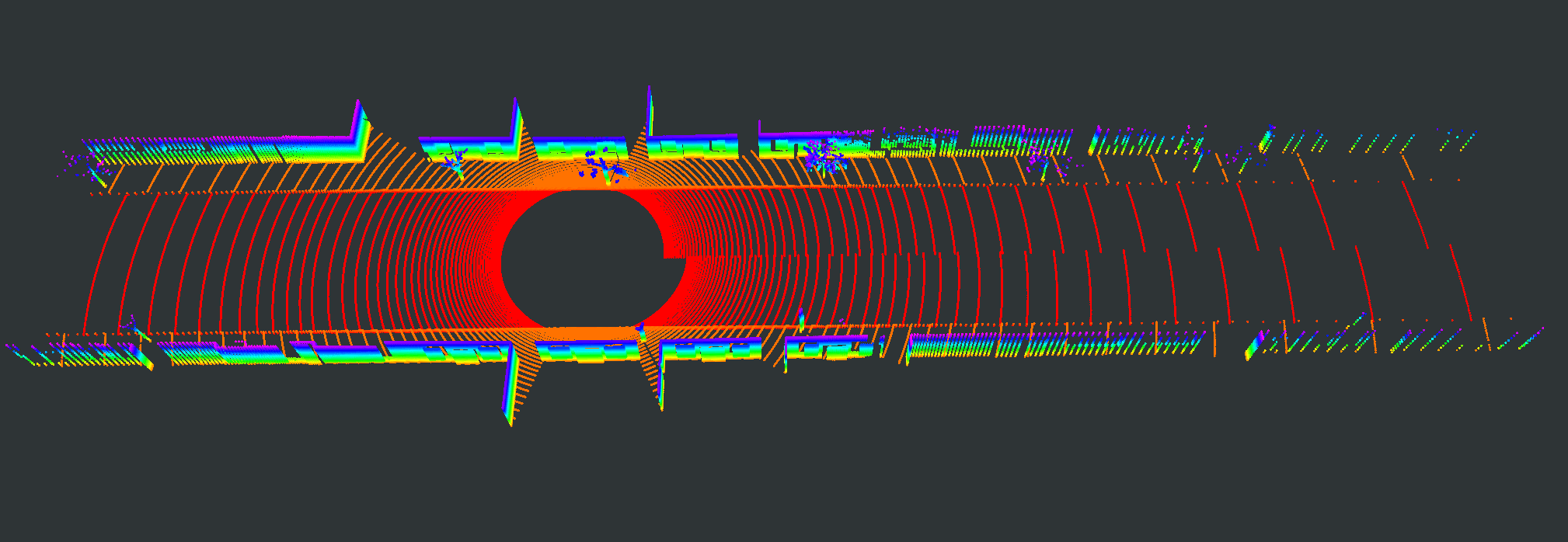}}} & \raisebox{-0.32\height}{\subfigure{\includegraphics[width=0.165\textwidth]{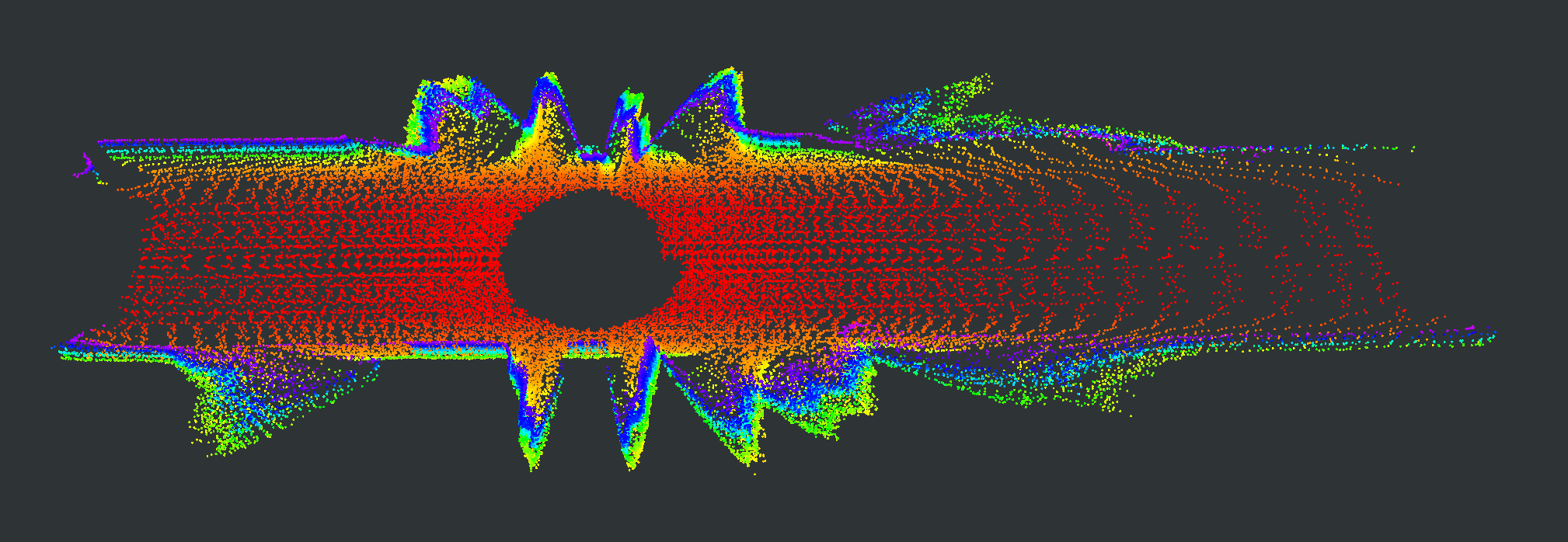}}} & \raisebox{-0.32\height}{\subfigure{\includegraphics[width=0.165\textwidth]{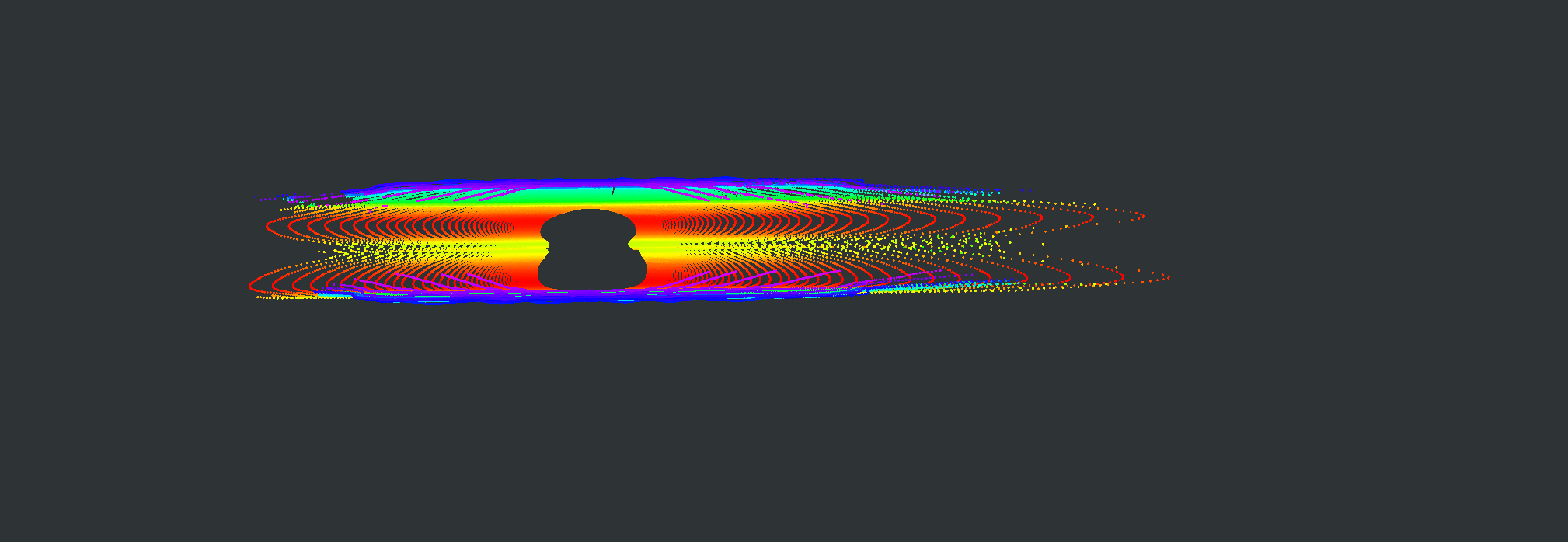}}} & \raisebox{-0.32\height}{\subfigure{\includegraphics[width=0.165\textwidth]{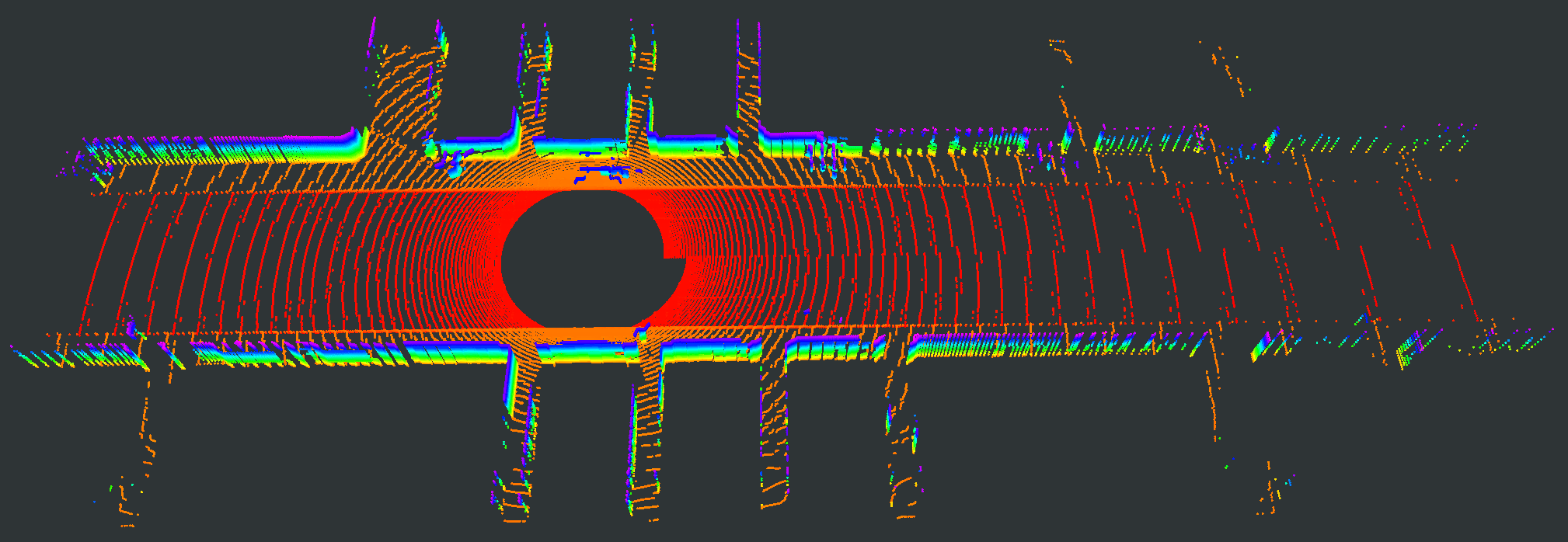}}}\\	
		\multicolumn{1}{c}{\fontsize{8}{8}\selectfont \begin{tabular}[c]{@{}c@{}}MaiCity-\\sequence01\\-0-49\end{tabular}} &\raisebox{-0.32\height}{\subfigure{\includegraphics[width=0.165\textwidth]{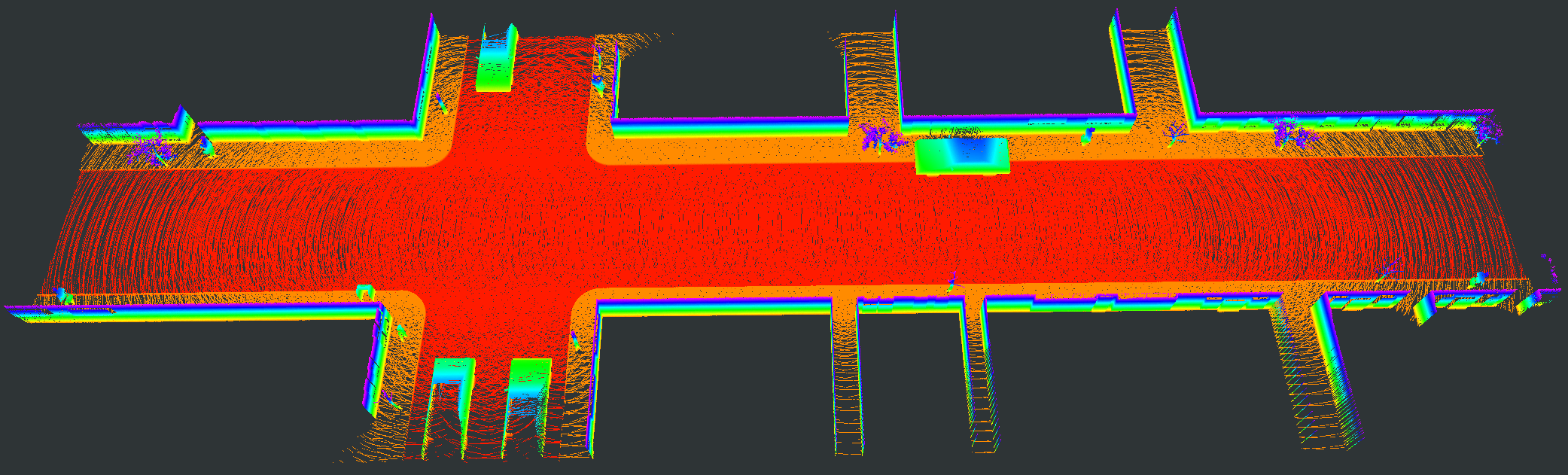}}} & \multicolumn{1}{c}{\fontsize{8}{8}\selectfont \begin{tabular}[c]{@{}c@{}}22\end{tabular}} & \raisebox{-0.32\height}{\subfigure{\includegraphics[width=0.165\textwidth]{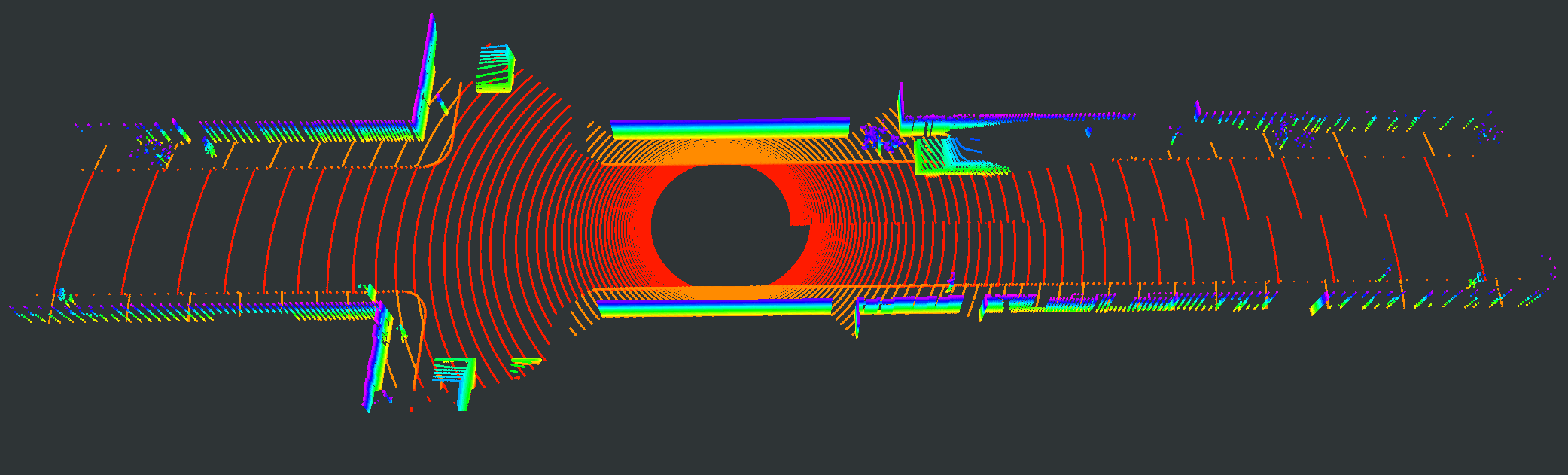}}} & \raisebox{-0.32\height}{\subfigure{\includegraphics[width=0.165\textwidth]{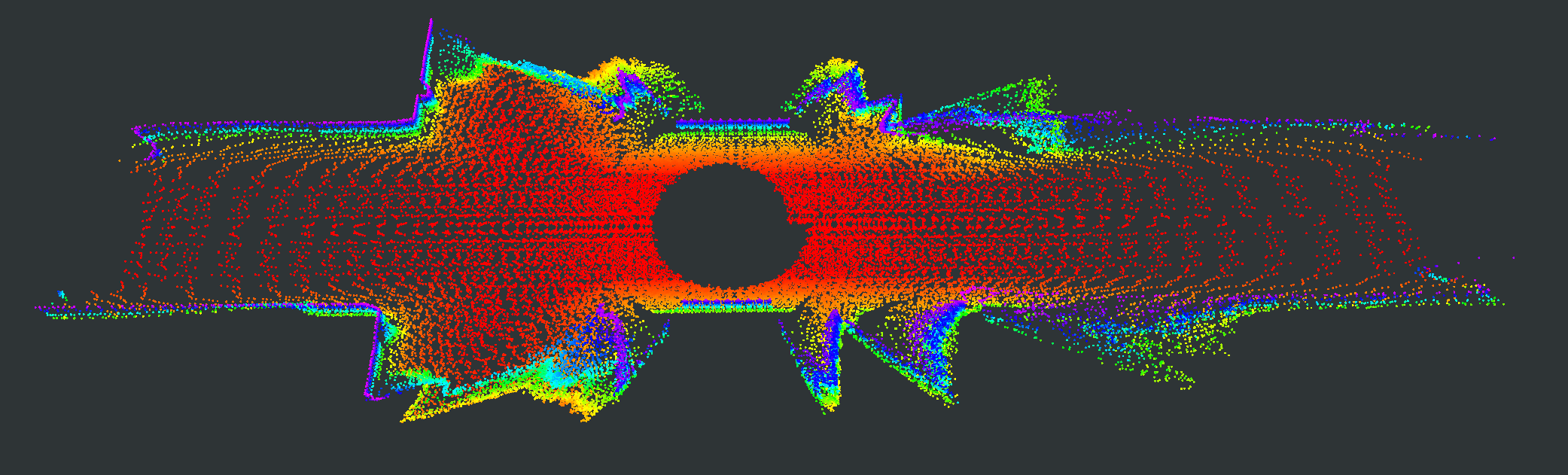}}} & \raisebox{-0.32\height}{\subfigure{\includegraphics[width=0.165\textwidth]{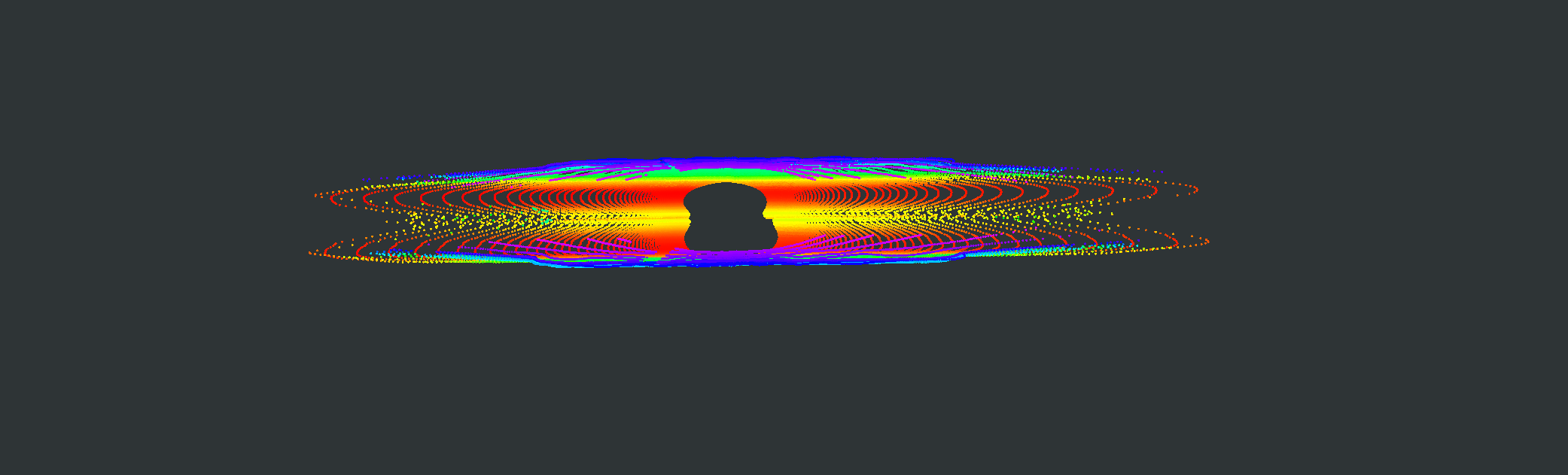}}} & \raisebox{-0.32\height}{\subfigure{\includegraphics[width=0.165\textwidth]{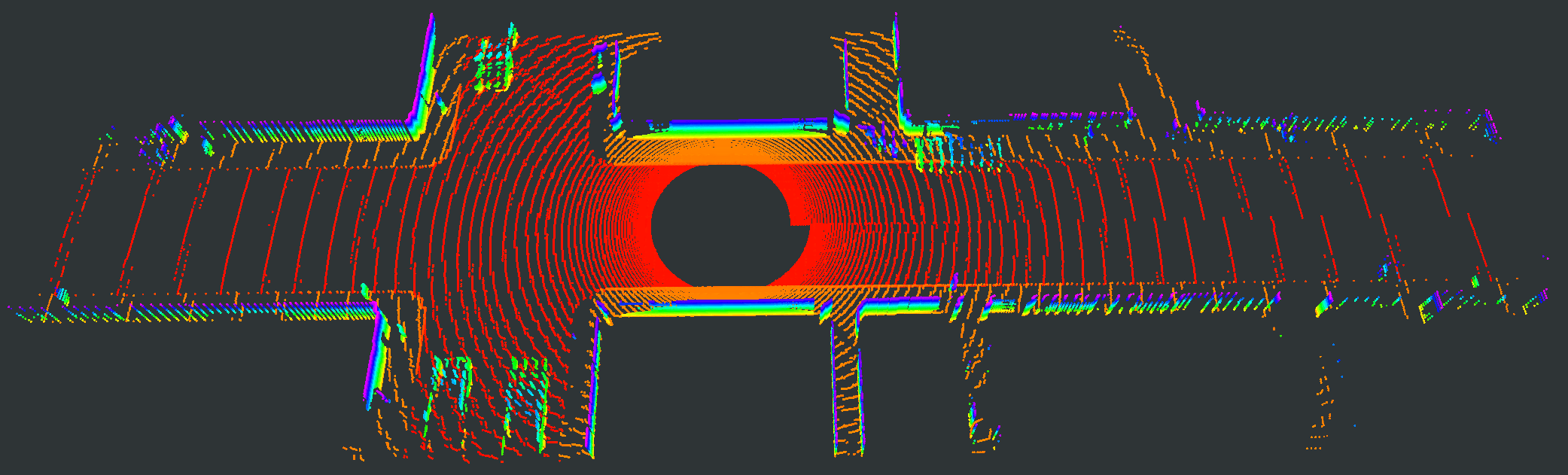}}}\\	
		\multicolumn{1}{c}{\fontsize{8}{8}\selectfont \begin{tabular}[c]{@{}c@{}}KITTI-\\sequence00\\-1151-1200\end{tabular}}
		&\raisebox{-0.38\height}{\subfigure{\includegraphics[width=0.165\textwidth]{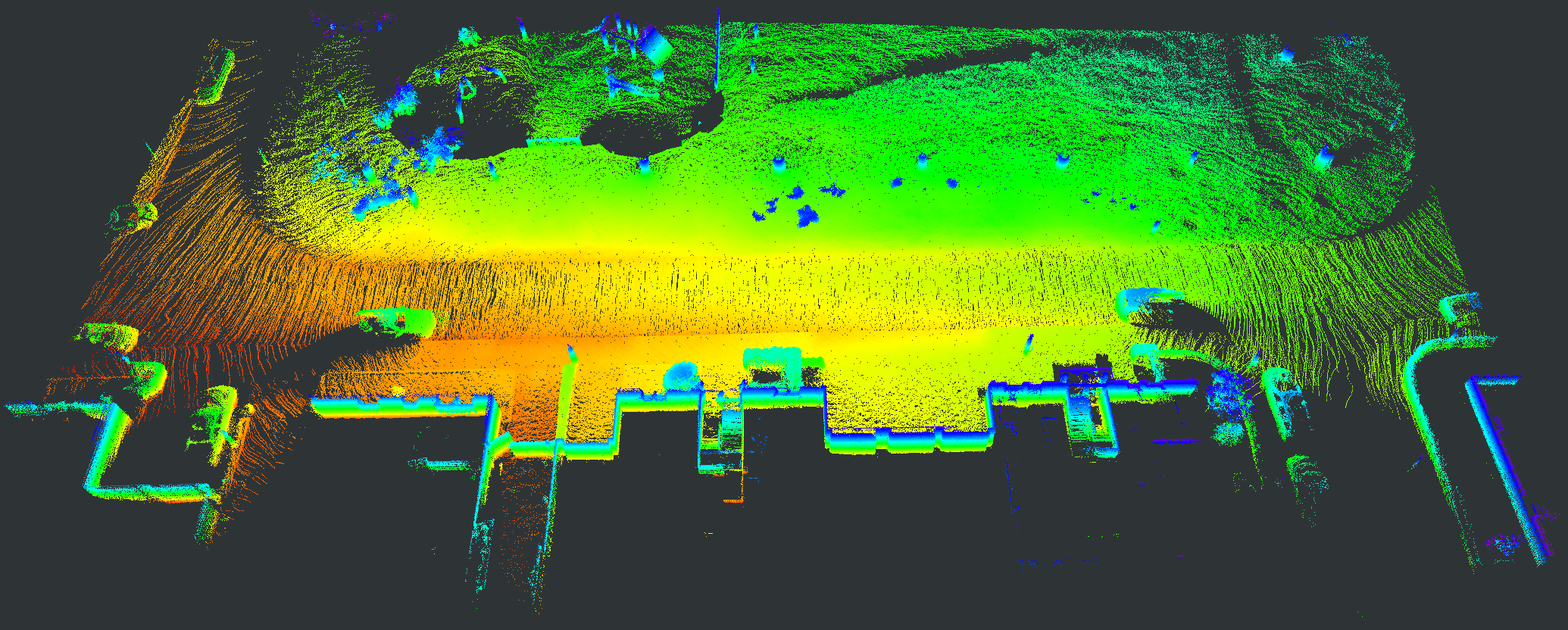}}} & \multicolumn{1}{c}{\fontsize{8}{8}\selectfont \begin{tabular}[c]{@{}c@{}}1173\end{tabular}} & \raisebox{-0.38\height}{\subfigure{\includegraphics[width=0.165\textwidth]{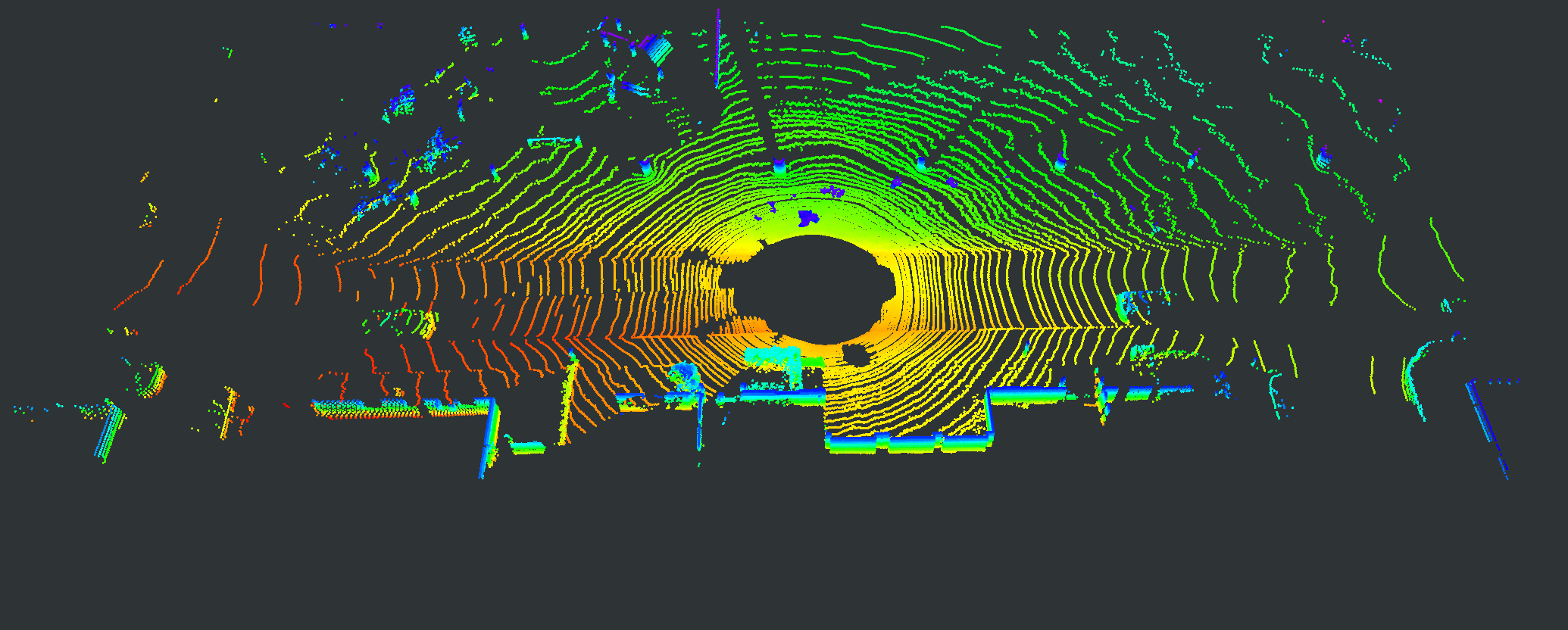}}} & \raisebox{-0.38\height}{\subfigure{\includegraphics[width=0.165\textwidth]{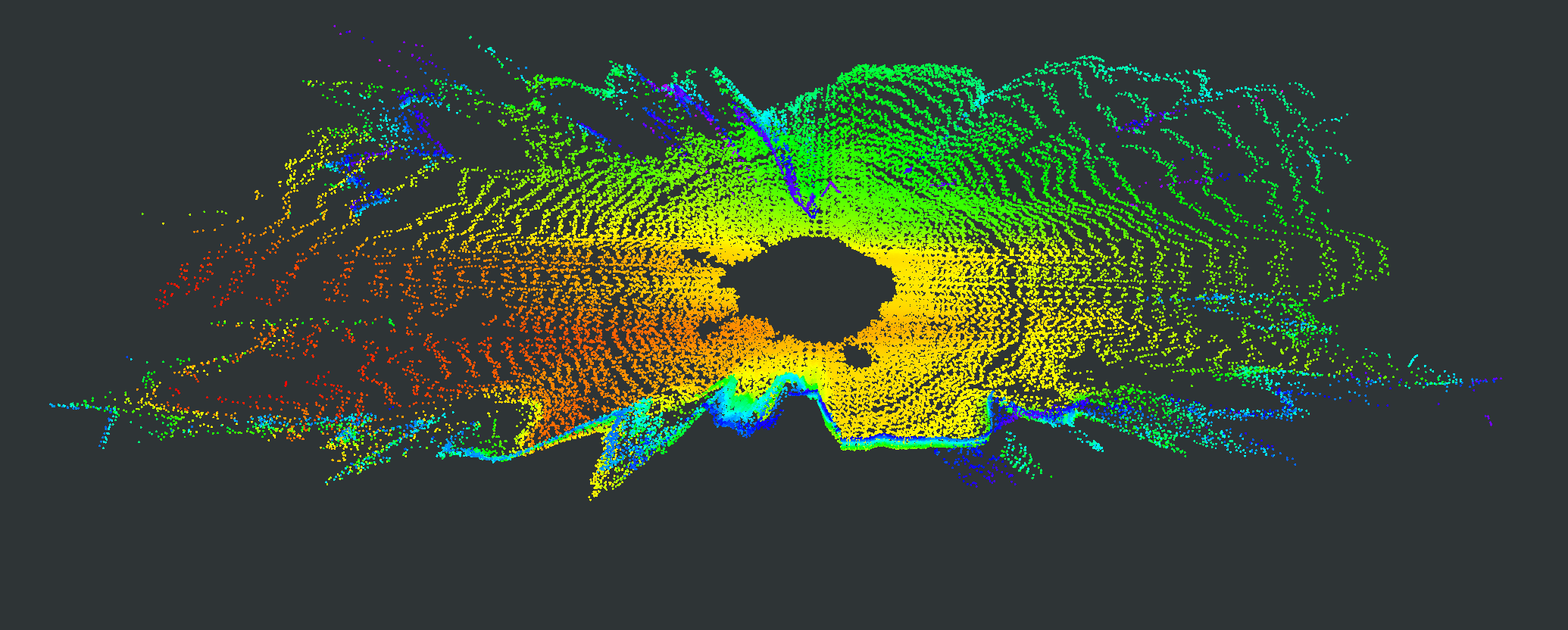}}} & \raisebox{-0.38\height}{\subfigure{\includegraphics[width=0.165\textwidth]{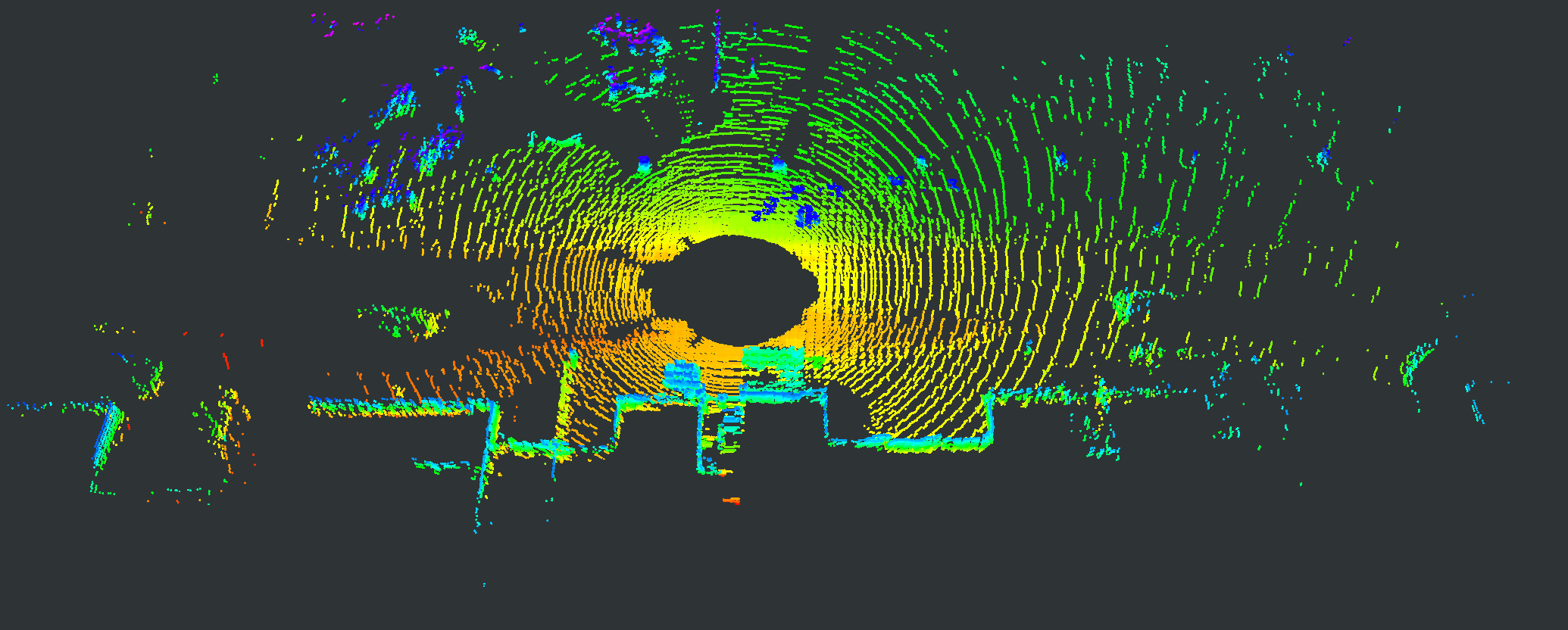}}} & \raisebox{-0.38\height}{\subfigure{\includegraphics[width=0.165\textwidth]{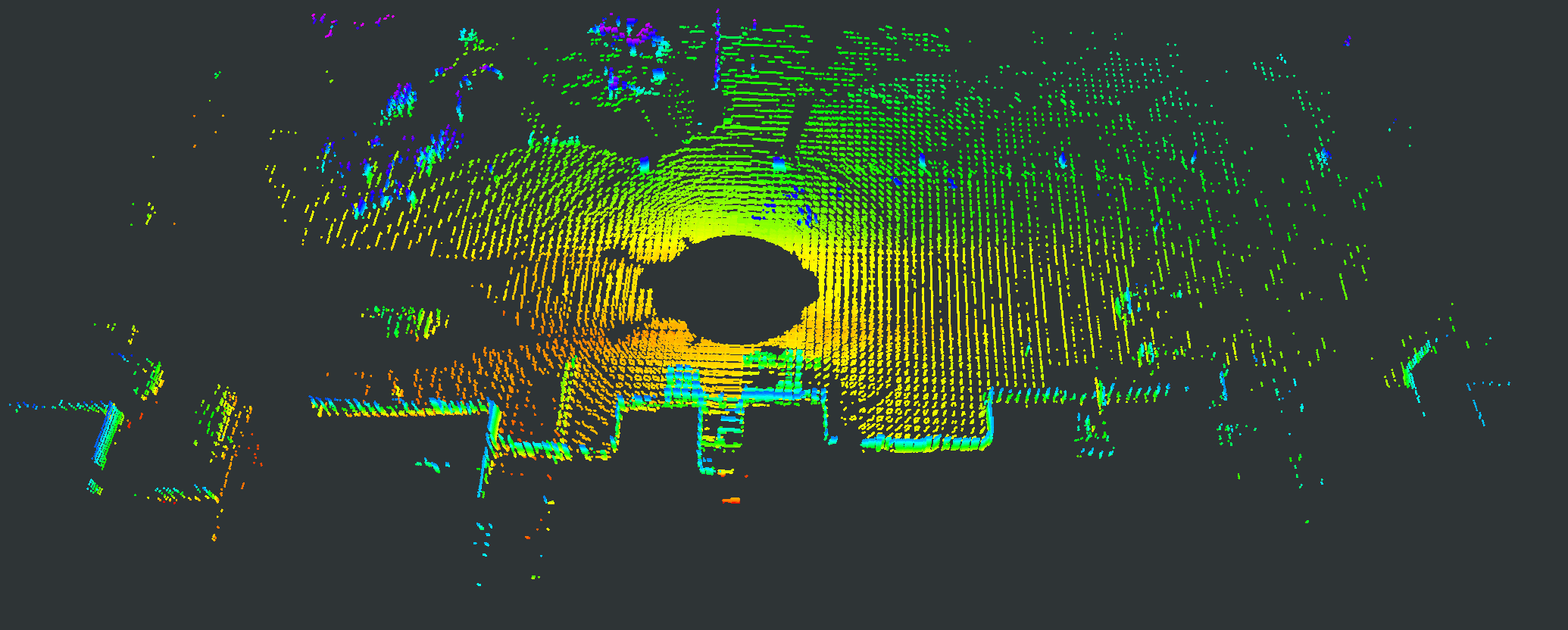}}}\\	
		\multicolumn{1}{c}{\fontsize{8}{8}\selectfont \begin{tabular}[c]{@{}c@{}}KITTI-\\sequence03\\-751-800\end{tabular}}
		&\raisebox{-0.38\height}{\subfigure{\includegraphics[width=0.165\textwidth]{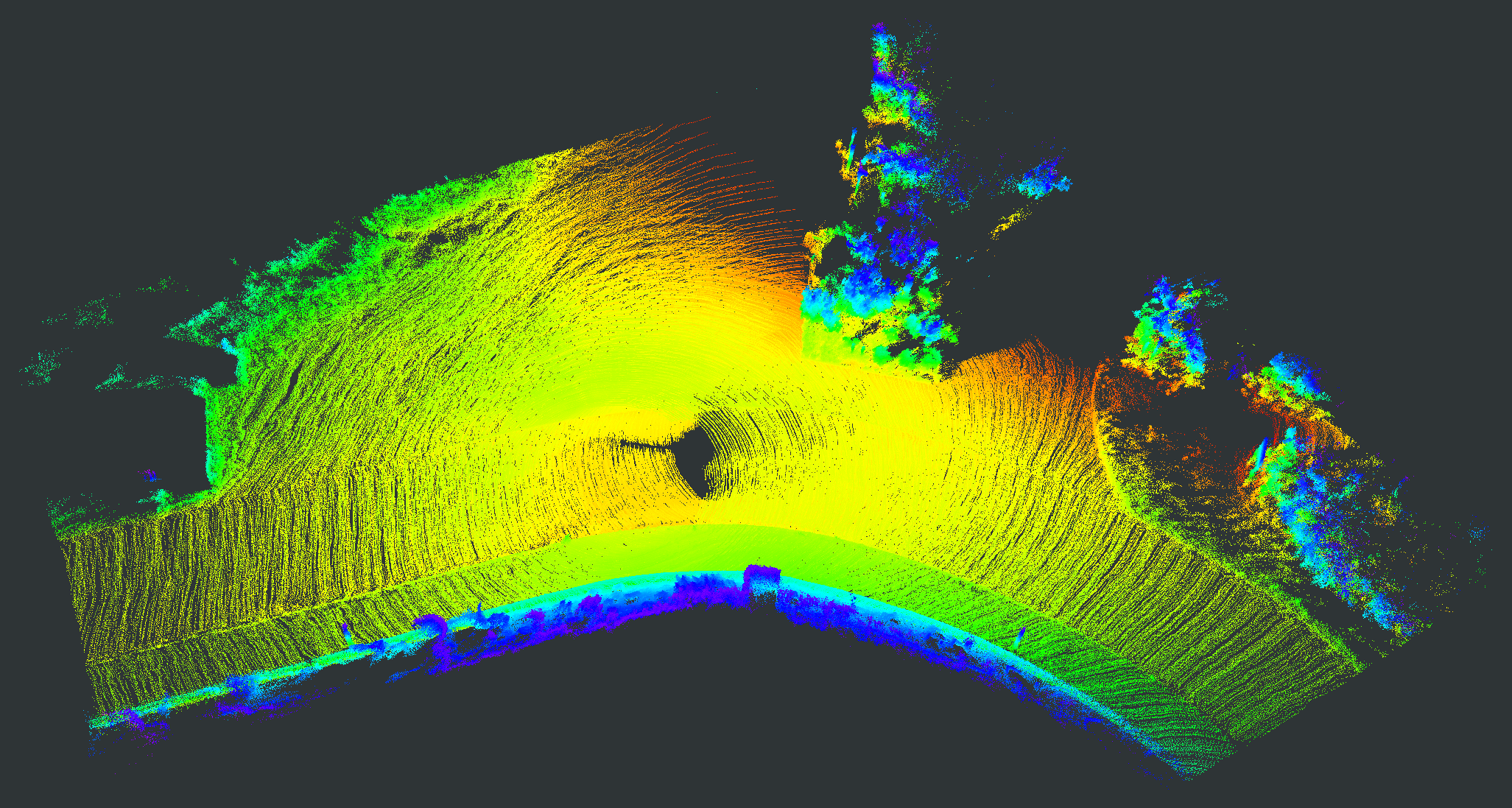}}} & \multicolumn{1}{c}{\fontsize{8}{8}\selectfont \begin{tabular}[c]{@{}c@{}}773\end{tabular}} & \raisebox{-0.38\height}{\subfigure{\includegraphics[width=0.165\textwidth]{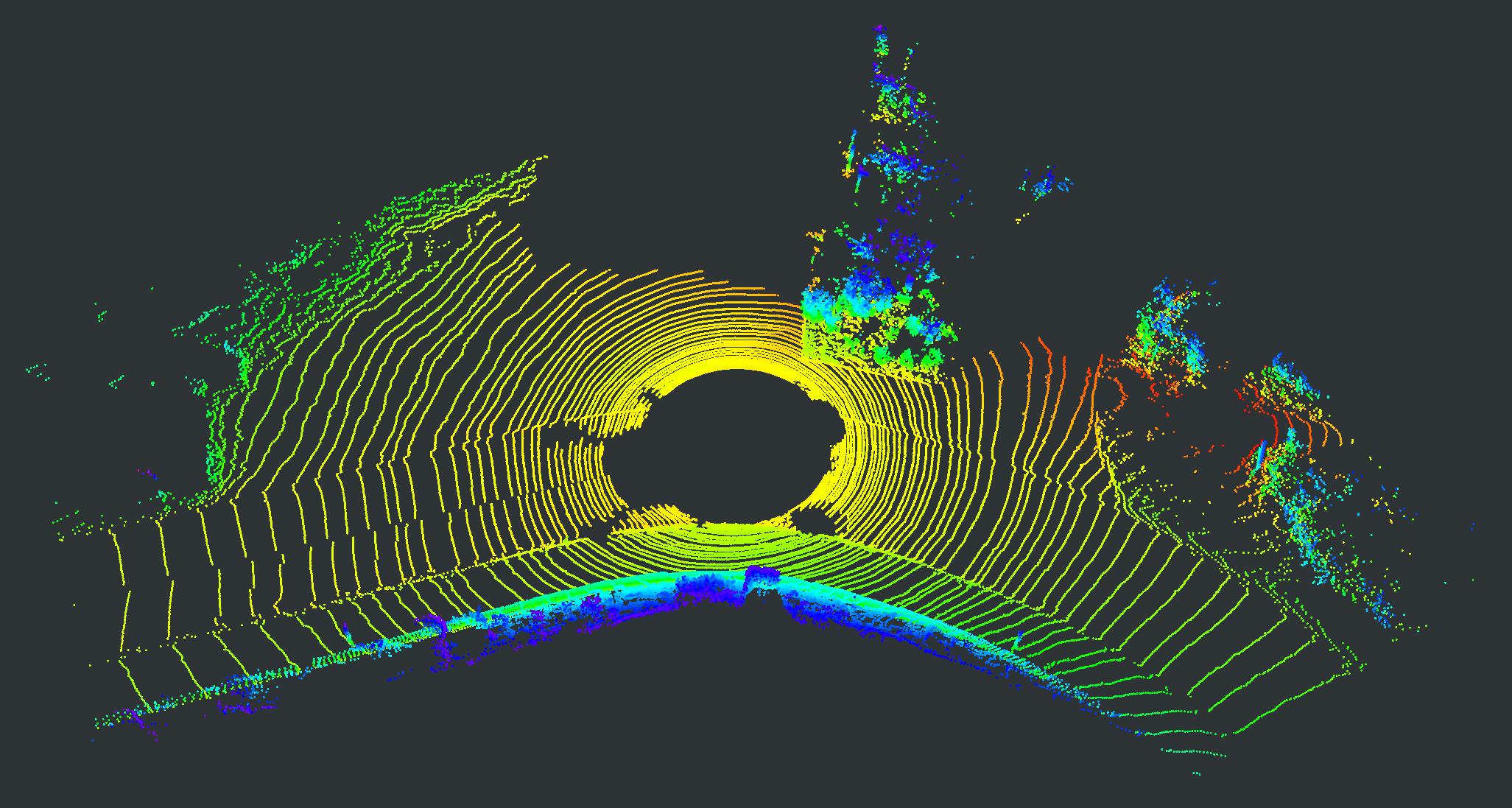}}} & \raisebox{-0.38\height}{\subfigure{\includegraphics[width=0.165\textwidth]{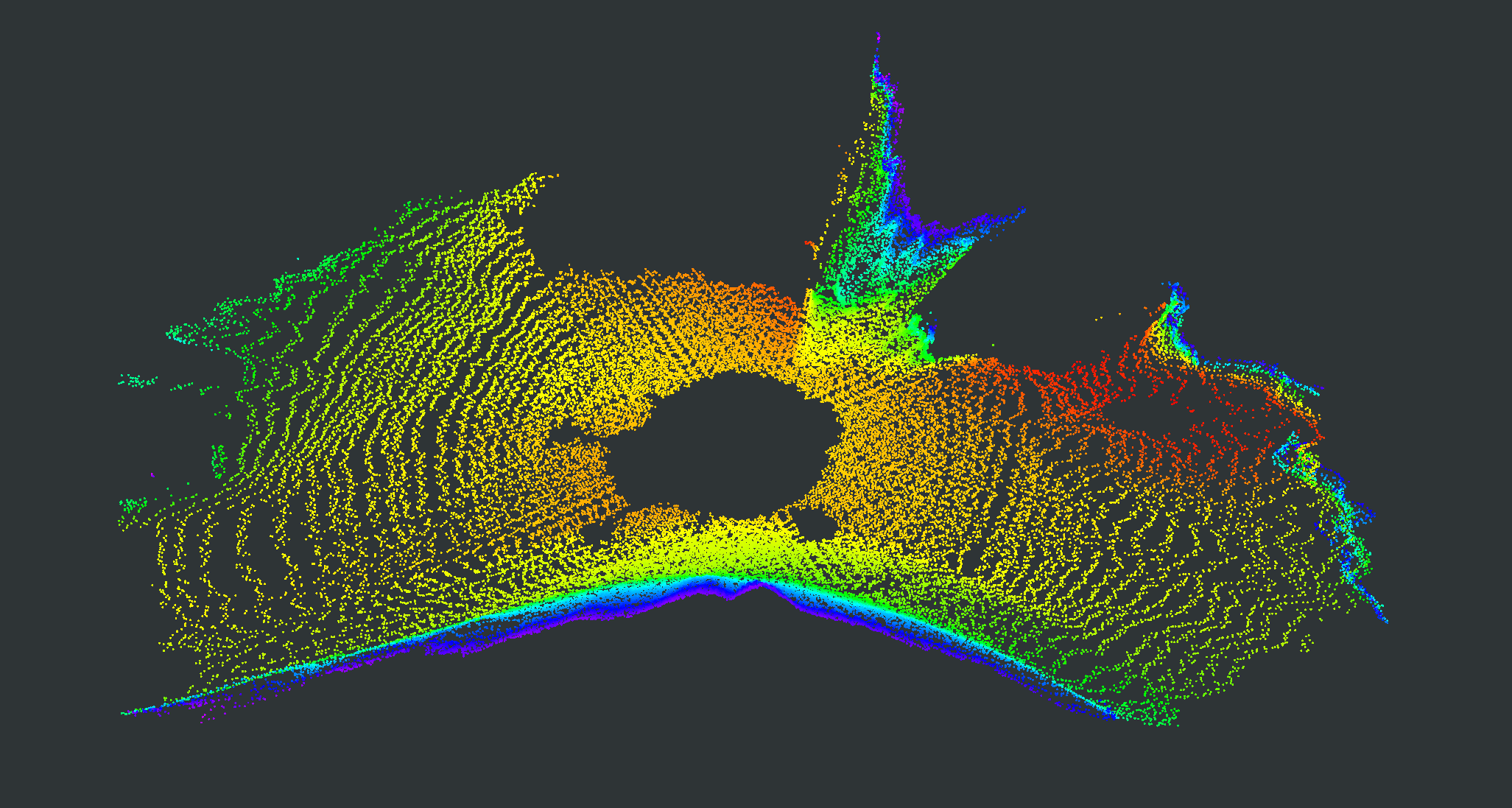}}} & \raisebox{-0.38\height}{\subfigure{\includegraphics[width=0.165\textwidth]{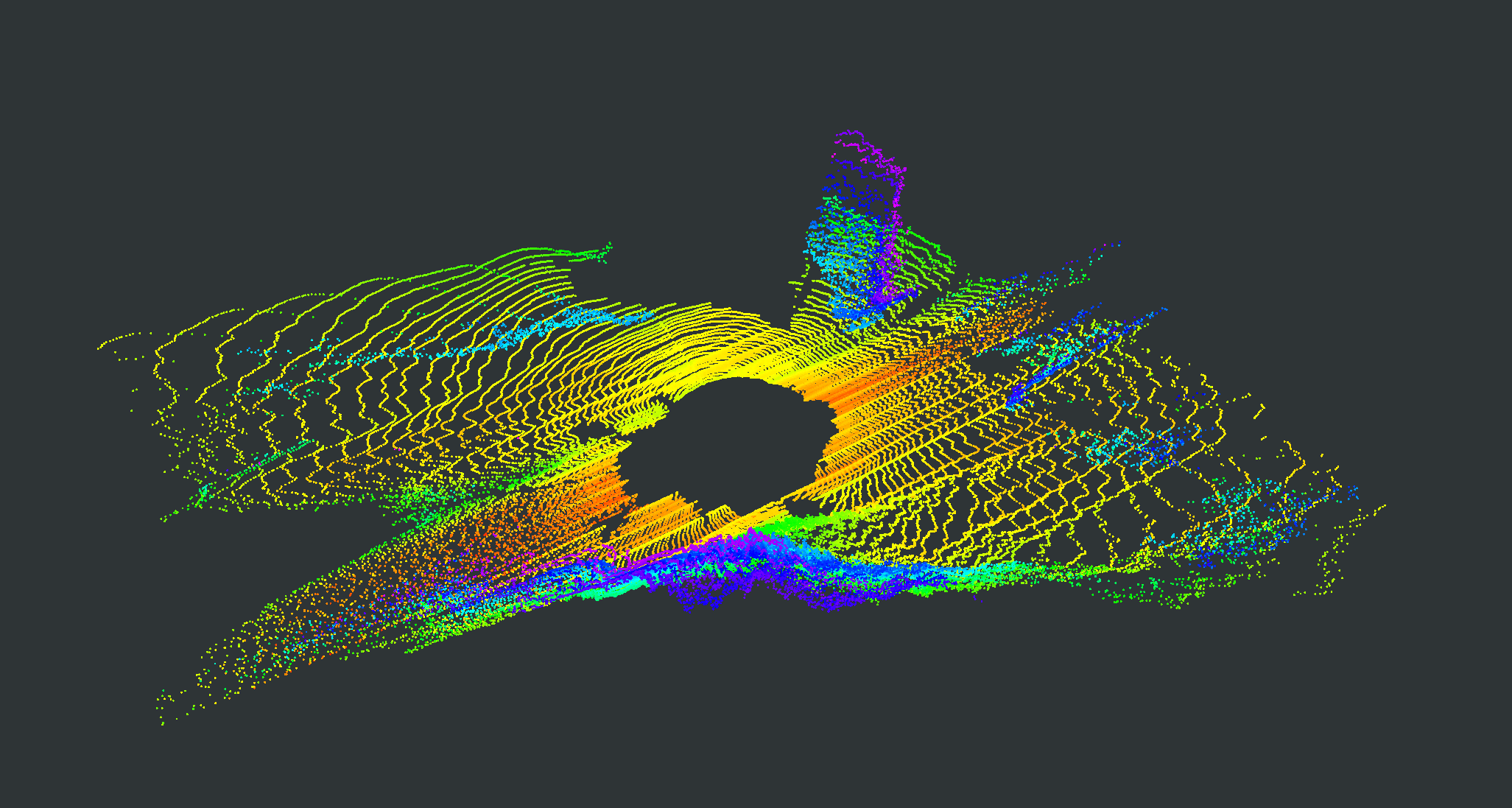}}} & \raisebox{-0.38\height}{\subfigure{\includegraphics[width=0.165\textwidth]{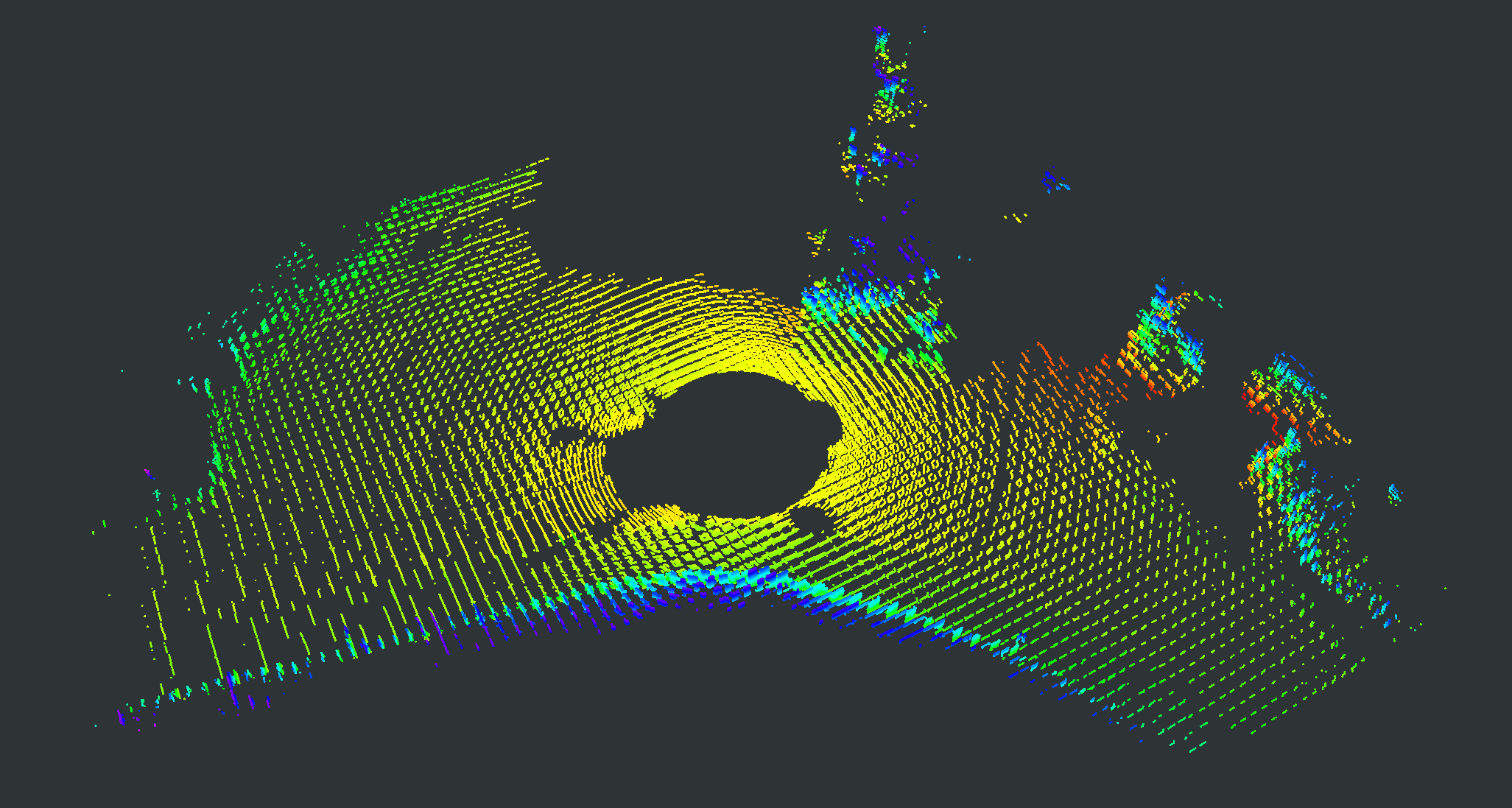}}}									
	\end{tabularx}	
	}
	\fontfamily{\rmdefault}\selectfont 
	\caption{Inference effects of three different methods on the MaiCity and KITTI datasets. MaiCity-sequence00-0-49 scene and MaiCity-sequence01-0-49 scene represent the first 50 scans of the MaiCity dataset 00 sequence and 01 sequence, respectively, while KITTI-sequence00-1151-1200 scene represents $1151\sim 1200$ scans of the KITTI 00 sequence and KITTI-sequence03-751-800 scene represents $751\sim 800$ scans of the KITTI 03 sequence. The last four columns show the inference effects corresponding to the 3rd column scans. The inference effects in this figure correspond to the green rows in Tab. \ref{tab:Parent-child NeRF Inference Effect}.}
	\label{Inference-effects}%
\end{figure*}

We evaluate our proposed PC-NeRF's inference accuracy and deployment potential in different scales qualitatively and quantitatively. We conduct experiments on the MaiCity and KITTI datasets using 50 consecutive scans as a scene, corresponding to the small-scale evaluation since each NeRF model is trained and evaluated only by one scene. As shown in Fig. \ref{Inference-effects}, MapRayCasting and OriginalNeRF can roughly reconstruct the environment while losing many environmental details. Especially when the input sensor data is distributed centrally over large spatial segments such as the ground, OriginalNeRF tends to be trained overfitting to the representation of larger objects and ignoring smaller objects, as shown in rows 2 and 3 of column 6 of Fig. \ref{Inference-effects}. In contrast, our proposed PC-NeRF achieves better results compared to the baselines by optimizing the scene-level, segment-level, and point-level environmental representations concurrently, as shown in the last column of Fig. \ref{Inference-effects}. We also report the quantitative results for small-scale scenes in Tab. \ref{tab:Parent-child NeRF Inference Effect}. As shown in Tab. \ref{tab:Parent-child NeRF Inference Effect}, our proposed PC-NeRF performs much better than MapRayCasting and OriginalNeRF and achieves a high novel LiDAR view synthesis accuracy (e.g. Avg. Error $<$ 0.50$\,\mathrm{m}$) and 3D reconstruction accuracy (e.g. CD $<$ 0.23$\,\mathrm{m}$). Moreover, our proposed two-step depth inference has much better inference accuracy than the one-step depth inference and is highly stable on our proposed PC-NeRF. Therefore, for the subsequent experiments in Sec.~\ref{sec:Lesser point clouds for 3D reconstruction} and Sec.~\ref{sec:Ablation}, we use the one-step depth inference and the two-step depth inference for OriginalNeRF and our proposed PC-NeRF, respectively. Compared to OriginalNeRF, which uses ten epochs of training, our proposed PC-NeRF demonstrates good deployment potential because our proposed PC-NeRF yields better results using only one training and does not change any model parameters for different scenes. 

\begin{figure}[!t]
	\centering
	\subfigure[Real LiDAR point clouds of all test scans.]{\includegraphics[width=3.5in]{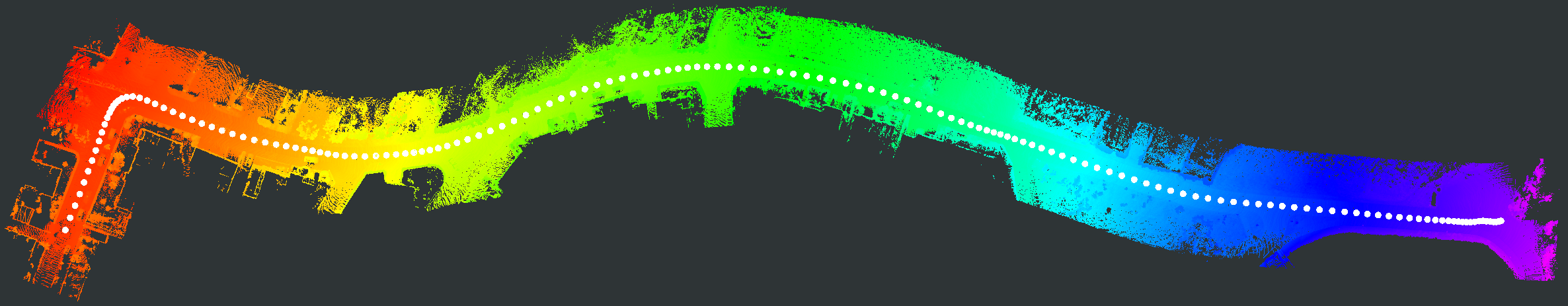}%
		\label{fig_source_merge}}
	\hspace{0.01in}		
	\subfigure[Reconstructed LiDAR point clouds by our proposed PC-NeRF.]{\includegraphics[width=3.5in]{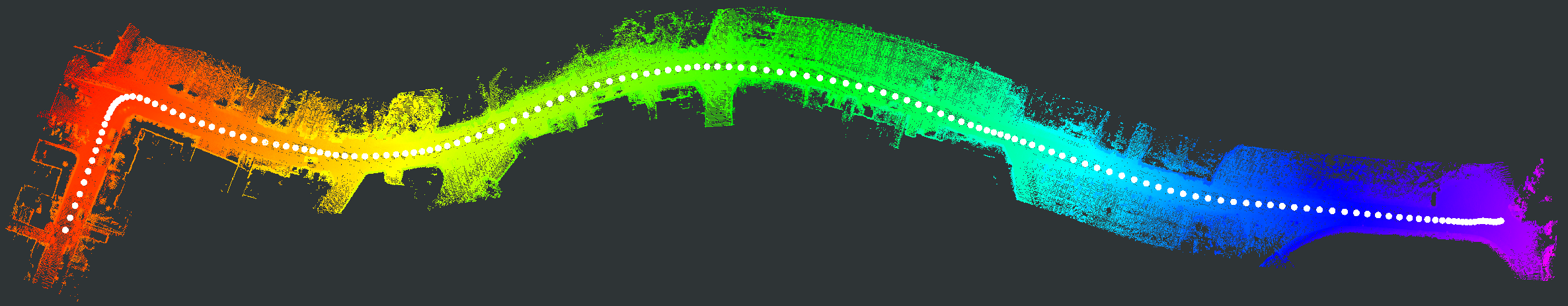}	
		\label{fig_prediction_pc_nerf_merge}}
	\hspace{0.01in}		
	\subfigure[Our proposed PC-NeRF 3D reconstruction accuracy.]{\includegraphics[width=3.5in]{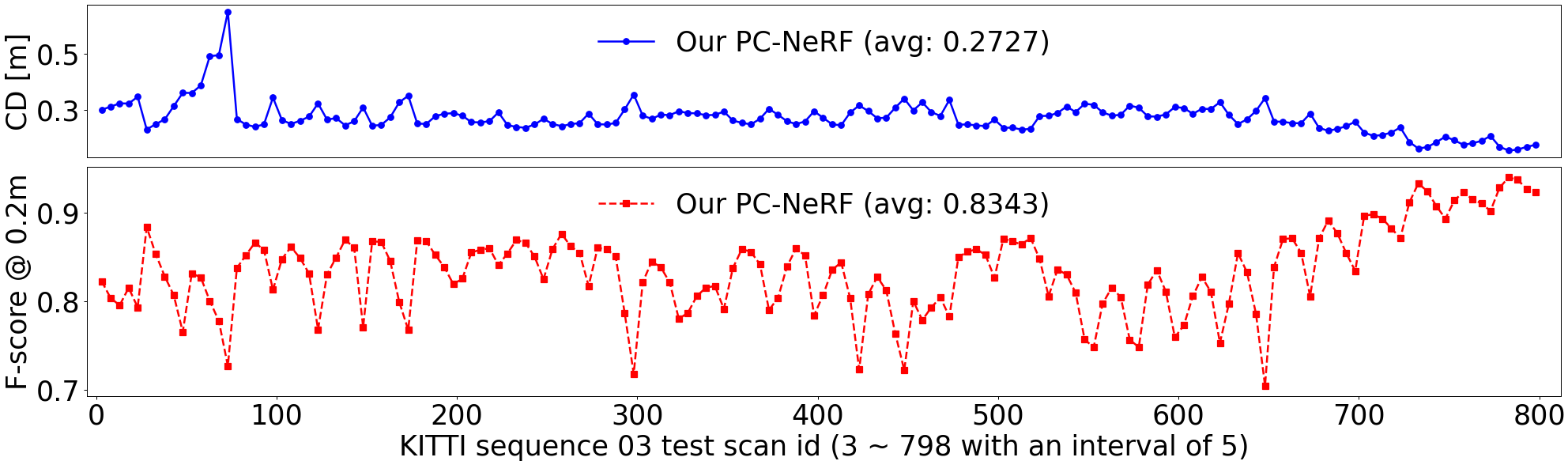}%
		\label{fig_cd_F_score2}}
	\hspace{0.01in}			
	\caption{KITTI 03 sequence 3D scene reconstruction using our proposed PC-NeRF. The white dots in Fig. \ref{fig_motivation}(a) and Fig. \ref{fig_motivation}(b) indicate the position of each scan, corresponding to each data point in Fig. \ref{fig_motivation}(c). As qualitative and quantitative results are shown in the three sub-figures, our proposed PC-NeRF has high 3D reconstruction accuracy in the KITTI 03 sequence.}
	\label{fig_motivation}
\end{figure}

To further explore the deployment potential, we test our proposed PC-NeRF on large-scale scenes. According to Sec.~\ref{sec:parent NeRF}, we divide the KITTI 03 sequence (800 scans, $555\times 120\times 7.2\,{\mathrm{m}}^{3}$) into 32 sequential scenes and each PC-NeRF model is trained and evaluated on one scene. The spatial extent of each scene is about $61.5\times 42.5\times 3\,{\mathrm{m}}^{3}$, overlapping with that of neighbouring scenes. In the 30th scene, our proposed PC-NeRF trains three epochs because one or two epochs of training result in the failure of two-step depth inference. Additionally, our proposed PC-NeRF trains only one epoch on all the remaining 31 scenes. As a comparison, OriginalNeRF trains three epochs on all 32 scenes. From Tab. \ref{tab:table-KITTI-sequence03} and Fig. \ref{fig_motivation}, it can be seen that our proposed PC-NeRF is trained to achieve a high novel LiDAR view synthesis accuracy (e.g., Avg. Error $<$ 0.66$\,\mathrm{m}$) and 3D reconstruction accuracy (e.g., CD $<$ 0.28$\,\mathrm{m}$), further demonstrating its promising deployment potential in large-scale environments.

\begin{table*}
	\centering
	\renewcommand\arraystretch{1.4}		
	\caption{Quantitative results for evaluation on small-scale scenes}
	{\fontsize{6.5}{8}\selectfont 
	\begin{tabularx}{\textwidth}{@{\hspace{0.0in}}>{\centering\arraybackslash}p{0.05in}@{\hspace{0.0in}}>{\centering\arraybackslash}X@{\hspace{0.0in}}>{\centering\arraybackslash}X@{\hspace{0.0in}}>{\centering\arraybackslash}p{0.05in}@{\hspace{0.0in}}>{\centering\arraybackslash}p{0.05in}@{\hspace{0.0in}}>{\centering\arraybackslash}p{0.05in}@{\hspace{0.0in}}>{\centering\arraybackslash}p{0.05in}@{\hspace{0.0in}}>{\centering\arraybackslash}p{0.05in}@{\hspace{0.0in}}>{\centering\arraybackslash}p{0.05in}|@{\hspace{0.0in}}>{\centering\arraybackslash}p{0.05in}@{\hspace{0.0in}}>{\centering\arraybackslash}p{0.05in}@{\hspace{0.0in}}>{\centering\arraybackslash}p{0.05in}@{\hspace{0.0in}}>{\centering\arraybackslash}p{0.05in}@{\hspace{0.0in}}>{\centering\arraybackslash}p{0.05in}@{\hspace{0.0in}}>{\centering\arraybackslash}p{0.05in}@{\hspace{0.0in}}}
	\cline{1-15}		
	\midrule	
	\multicolumn{1}{c}{\begin{tabular}[c]{@{}c@{}}Method\end{tabular}} & \multicolumn{1}{c}{\begin{tabular}[c]{@{}c@{}}Depth\\inference \end{tabular}} & \multicolumn{1}{c}{\begin{tabular}[c]{@{}c@{}}Train\\epoch \end{tabular}} & \multicolumn{1}{|c}{\begin{tabular}[c]{@{}c@{}}Avg.\\Error{[}$\mathrm{m}${]}\end{tabular}} & \multicolumn{1}{c}{\begin{tabular}[c]{@{}c@{}}Acc@\\0.2$\mathrm{m}${[}\%{]}\end{tabular}} & \multicolumn{1}{c}{\begin{tabular}[c]{@{}c@{}}Acc@\\1$\mathrm{m}${[}\%{]}\end{tabular}}
	& \multicolumn{1}{c}{\begin{tabular}[c]{@{}c@{}}CD\\{[}$\mathrm{m}${]}\end{tabular}} & \multicolumn{1}{c}{\begin{tabular}[c]{@{}c@{}}F-score\\@0.2$\mathrm{m}$\end{tabular}} & \multicolumn{1}{c|}{\begin{tabular}[c]{@{}c@{}}F-score\\@1$\mathrm{m}$\end{tabular}}
	& \multicolumn{1}{c}{\begin{tabular}[c]{@{}c@{}}Avg.\\Error{[}$\mathrm{m}${]}\end{tabular}} & \multicolumn{1}{c}{\begin{tabular}[c]{@{}c@{}}Acc@\\0.2$\mathrm{m}${[}\%{]}\end{tabular}} & \multicolumn{1}{c}{\begin{tabular}[c]{@{}c@{}}Acc@\\1$\mathrm{m}${[}\%{]}\end{tabular}}
	& \multicolumn{1}{c}{\begin{tabular}[c]{@{}c@{}}CD\\{[}$\mathrm{m}${]}\end{tabular}} & \multicolumn{1}{c}{\begin{tabular}[c]{@{}c@{}}F-score\\@0.2$\mathrm{m}$\end{tabular}} & \multicolumn{1}{c}{\begin{tabular}[c]{@{}c@{}}F-score\\@1$\mathrm{m}$\end{tabular}}\\
	\cline{1-15}	
	\midrule		    
	&    &   & \multicolumn{6}{|c|}{\textbf{MaiCity-sequence00-0-49}}     & \multicolumn{6}{c}{\textbf{MaiCity-sequence01-0-49}} \\
	\hline
	\multicolumn{1}{c}{MapRayCasting} & \multicolumn{1}{c}{-} & \multicolumn{1}{c}{-} & \multicolumn{1}{|c}{0.841} & \multicolumn{1}{c}{49.873} & \multicolumn{1}{c}{78.040} & \multicolumn{1}{c}{0.600} & \multicolumn{1}{c}{0.736} & \multicolumn{1}{c|}{0.917} & \multicolumn{1}{c}{0.744} & \multicolumn{1}{c}{55.836} & \multicolumn{1}{c}{80.781} & \multicolumn{1}{c}{0.505} & \multicolumn{1}{c}{0.754} & \multicolumn{1}{c}{0.943} \\
	\multicolumn{1}{c}{\cellcolor[HTML]{34FF34}OriginalNeRF} &  
	\multicolumn{1}{c}{\cellcolor[HTML]{34FF34}one-step}
	 &\multicolumn{1}{c}{\cellcolor[HTML]{34FF34}10} & \multicolumn{1}{|c}{\cellcolor[HTML]{34FF34}3.357} & \multicolumn{1}{c}{\cellcolor[HTML]{34FF34}0.004} & \multicolumn{1}{c}{\cellcolor[HTML]{34FF34}0.247} & \multicolumn{1}{c}{\cellcolor[HTML]{34FF34}3.184} & \multicolumn{1}{c}{\cellcolor[HTML]{34FF34}NaN}   & \multicolumn{1}{c|}{\cellcolor[HTML]{34FF34}0.224} & \multicolumn{1}{c}{\cellcolor[HTML]{34FF34}3.699} & \multicolumn{1}{c}{\cellcolor[HTML]{34FF34}0.312} & \multicolumn{1}{c}{\cellcolor[HTML]{34FF34}2.009} & \multicolumn{1}{c}{\cellcolor[HTML]{34FF34}3.399} & \multicolumn{1}{c}{\cellcolor[HTML]{34FF34}NaN}   & \multicolumn{1}{c}{\cellcolor[HTML]{34FF34}0.249} \\
	 \multicolumn{1}{c}{\cellcolor[HTML]{FFCCC9}OriginalNeRF} &  \multicolumn{1}{c}{\cellcolor[HTML]{FFCCC9}two-step} 
	 & \multicolumn{1}{c}{\cellcolor[HTML]{FFCCC9}10} & \multicolumn{1}{|c}{\cellcolor[HTML]{FFCCC9}0.588} & \multicolumn{1}{c}{\cellcolor[HTML]{FFCCC9}83.991} & \multicolumn{1}{c}{\cellcolor[HTML]{FFCCC9}87.956} & \multicolumn{1}{c}{\cellcolor[HTML]{FFCCC9}0.315} & \multicolumn{1}{c}{\cellcolor[HTML]{FFCCC9}0.918} & \multicolumn{1}{c|}{\cellcolor[HTML]{FFCCC9}0.962} & \multicolumn{1}{c}{\cellcolor[HTML]{FFCCC9}0.459} & \multicolumn{1}{c}{\cellcolor[HTML]{FFCCC9}85.548} & \multicolumn{1}{c}{\cellcolor[HTML]{FFCCC9}91.387} & \multicolumn{1}{c}{\cellcolor[HTML]{FFCCC9}0.233} & \multicolumn{1}{c}{\cellcolor[HTML]{FFCCC9}0.919} & \multicolumn{1}{c}{\cellcolor[HTML]{FFCCC9}0.975} \\
	\multicolumn{1}{c}{\cellcolor[HTML]{FFCCC9}PC-NeRF} & \multicolumn{1}{c}{\cellcolor[HTML]{FFCCC9}one-step} & \multicolumn{1}{c}{\cellcolor[HTML]{FFCCC9}1} & \multicolumn{1}{|c}{\cellcolor[HTML]{FFCCC9}1.755} & \multicolumn{1}{c}{\cellcolor[HTML]{FFCCC9}2.575} & \multicolumn{1}{c}{\cellcolor[HTML]{FFCCC9}23.879} & \multicolumn{1}{c}{\cellcolor[HTML]{FFCCC9}1.279} & \multicolumn{1}{c}{\cellcolor[HTML]{FFCCC9}0.070} & \multicolumn{1}{c|}{\cellcolor[HTML]{FFCCC9}0.868} & \multicolumn{1}{c}{\cellcolor[HTML]{FFCCC9}2.025} & \multicolumn{1}{c}{\cellcolor[HTML]{FFCCC9}2.189} & \multicolumn{1}{c}{\cellcolor[HTML]{FFCCC9}16.672} & \multicolumn{1}{c}{\cellcolor[HTML]{FFCCC9}1.455} & \multicolumn{1}{c}{\cellcolor[HTML]{FFCCC9}0.056} & \multicolumn{1}{c}{\cellcolor[HTML]{FFCCC9}0.808} \\
	\multicolumn{1}{c}{\textbf{\cellcolor[HTML]{34FF34}PC-NeRF}} & \multicolumn{1}{c}{\textbf{\cellcolor[HTML]{34FF34}two-step}} &\multicolumn{1}{c}{\cellcolor[HTML]{34FF34}1} & \multicolumn{1}{|c}{\textbf{\cellcolor[HTML]{34FF34}0.303}} & \multicolumn{1}{c}{\textbf{\cellcolor[HTML]{34FF34}88.956}} & \multicolumn{1}{c}{\textbf{\cellcolor[HTML]{34FF34}93.579}} & \multicolumn{1}{c}{\textbf{\cellcolor[HTML]{34FF34}0.172}} & \multicolumn{1}{c}{\textbf{\cellcolor[HTML]{34FF34}0.955}} & \multicolumn{1}{c|}{\textbf{\cellcolor[HTML]{34FF34}0.985}} & \multicolumn{1}{c}{\textbf{\cellcolor[HTML]{34FF34}0.392}} & \multicolumn{1}{c}{\textbf{\cellcolor[HTML]{34FF34}86.798}} & \multicolumn{1}{c}{\textbf{\cellcolor[HTML]{34FF34}92.568}} & \multicolumn{1}{c}{\textbf{\cellcolor[HTML]{34FF34}0.185}} & \multicolumn{1}{c}{\textbf{\cellcolor[HTML]{34FF34}0.935}} & \multicolumn{1}{c}{\textbf{\cellcolor[HTML]{34FF34}0.981}} \\
	\cline{1-15}	
	\midrule			
	& & &\multicolumn{6}{|c|}{\textbf{KITTI-sequence00-1151-1200}} & \multicolumn{6}{c}{\textbf{KITTI-sequence03-751-800}} \\
	\hline		
	\multicolumn{1}{c}{MapRayCasting} & \multicolumn{1}{c}{-} & \multicolumn{1}{c}{-} & \multicolumn{1}{|c}{0.871} & \multicolumn{1}{c}{44.358} & \multicolumn{1}{c}{77.823} & \multicolumn{1}{c}{0.466} & \multicolumn{1}{c}{0.740} & \multicolumn{1}{c|}{0.949} & \multicolumn{1}{c}{0.472} & \multicolumn{1}{c}{46.988} & \multicolumn{1}{c}{36.969} & \multicolumn{1}{c}{0.274} & \multicolumn{1}{c}{0.836} & \multicolumn{1}{c}{0.798} \\
	\multicolumn{1}{c}{\cellcolor[HTML]{34FF34}OriginalNeRF} &  \multicolumn{1}{c}{\cellcolor[HTML]{34FF34}one-step} & \multicolumn{1}{c}{\cellcolor[HTML]{34FF34}10} & \multicolumn{1}{|c}{\cellcolor[HTML]{34FF34}5.251} & \multicolumn{1}{c}{\cellcolor[HTML]{34FF34}0.261} & \multicolumn{1}{c}{\cellcolor[HTML]{34FF34}1.515} & \multicolumn{1}{c}{\cellcolor[HTML]{34FF34}3.289} & \multicolumn{1}{c}{\cellcolor[HTML]{34FF34}0.012} & \multicolumn{1}{c|}{\cellcolor[HTML]{34FF34}0.321} & \multicolumn{1}{c}{\cellcolor[HTML]{34FF34}1.590} & \multicolumn{1}{c}{\cellcolor[HTML]{34FF34}8.887} & \multicolumn{1}{c}{\cellcolor[HTML]{34FF34}53.757} & \multicolumn{1}{c}{\cellcolor[HTML]{34FF34}0.850} & \multicolumn{1}{c}{\cellcolor[HTML]{34FF34}0.328} & \multicolumn{1}{c}{\cellcolor[HTML]{34FF34}0.924} \\
	\multicolumn{1}{c}{\cellcolor[HTML]{FFCCC9}OriginalNeRF} & \multicolumn{1}{c}{\cellcolor[HTML]{FFCCC9}two-step} & \multicolumn{1}{c}{\cellcolor[HTML]{FFCCC9}10} & \multicolumn{1}{|c}{\cellcolor[HTML]{FFCCC9}0.515} & \multicolumn{1}{c}{\cellcolor[HTML]{FFCCC9}64.566} & \multicolumn{1}{c}{\cellcolor[HTML]{FFCCC9}91.948} & \multicolumn{1}{c}{\textbf{\cellcolor[HTML]{FFCCC9}0.223}} & \multicolumn{1}{c}{\cellcolor[HTML]{FFCCC9}0.888} & \multicolumn{1}{c|}{\textbf{\cellcolor[HTML]{FFCCC9}0.994}} & \multicolumn{6}{c}{\cellcolor[HTML]{FFCCC9}infer error} \\	
	\multicolumn{1}{c}{\cellcolor[HTML]{FFCCC9}PC-NeRF} & \multicolumn{1}{c}{\cellcolor[HTML]{FFCCC9}one-step} &  \multicolumn{1}{c}{\cellcolor[HTML]{FFCCC9}1}     & \multicolumn{1}{|c}{\cellcolor[HTML]{FFCCC9}2.331} & \multicolumn{1}{c}{\cellcolor[HTML]{FFCCC9}5.790} & \multicolumn{1}{c}{\cellcolor[HTML]{FFCCC9}27.583} & \multicolumn{1}{c}{\cellcolor[HTML]{FFCCC9}1.620} & \multicolumn{1}{c}{\cellcolor[HTML]{FFCCC9}0.178} & \multicolumn{1}{c|}{\cellcolor[HTML]{FFCCC9}0.836} & \multicolumn{1}{c}{\cellcolor[HTML]{FFCCC9}5.116} & \multicolumn{1}{c}{\cellcolor[HTML]{FFCCC9}0.482} & \multicolumn{1}{c}{\cellcolor[HTML]{FFCCC9}2.483} & \multicolumn{1}{c}{\cellcolor[HTML]{FFCCC9}4.192} & \multicolumn{1}{c}{\cellcolor[HTML]{FFCCC9}0.022} & \multicolumn{1}{c}{\cellcolor[HTML]{FFCCC9}0.236} \\	
	\multicolumn{1}{c}{\textbf{\cellcolor[HTML]{34FF34}PC-NeRF}}& \multicolumn{1}{c}{\textbf{\cellcolor[HTML]{34FF34}two-step}}&
	\multicolumn{1}{c}{\cellcolor[HTML]{34FF34}1} & \multicolumn{1}{|c}{\textbf{\cellcolor[HTML]{34FF34}0.488}} & \multicolumn{1}{c}{\textbf{\cellcolor[HTML]{34FF34}66.654}} & \multicolumn{1}{c}{\textbf{\cellcolor[HTML]{34FF34}92.131}} & \multicolumn{1}{c}{\cellcolor[HTML]{34FF34}0.224} & \multicolumn{1}{c}{\textbf{\cellcolor[HTML]{34FF34}0.891}} & \multicolumn{1}{c|}{\cellcolor[HTML]{34FF34}0.993} & \multicolumn{1}{c}{\textbf{\cellcolor[HTML]{34FF34}0.397}} & \multicolumn{1}{c}{\textbf{\cellcolor[HTML]{34FF34}51.264}} & \multicolumn{1}{c}{\textbf{\cellcolor[HTML]{34FF34}93.495}} & \multicolumn{1}{c}{\textbf{\cellcolor[HTML]{34FF34}0.224}} & \multicolumn{1}{c}{\textbf{\cellcolor[HTML]{34FF34}0.878}} & \multicolumn{1}{c}{\textbf{\cellcolor[HTML]{34FF34}0.999}} \\
	\cline{1-15}		
	\midrule	
	\end{tabularx}%
	}
	\label{tab:Parent-child NeRF Inference Effect}%
\end{table*}%

\begin{table}[!t]
	\renewcommand\arraystretch{1.4}	
	\caption{Quantitative results for evaluation on large-scale scenes}	
	{\fontsize{7.5}{8}\selectfont 
	\begin{tabularx}{\columnwidth}{>{\centering\arraybackslash}p{0.1in}>{\centering\arraybackslash}p{0.1in}>{\centering\arraybackslash}p{0.1in}>{\centering\arraybackslash}p{0.1in}>{\centering\arraybackslash}p{0.1in}>{\centering\arraybackslash}p{0.1in}}
		\cline{1-6}	
		\multicolumn{1}{c}{Method}        
		&\multicolumn{1}{c}{\begin{tabular}[c]{@{}c@{}}Depth\\inference\end{tabular}}
		&\multicolumn{1}{c}{\begin{tabular}[c]{@{}c@{}}Avg.\\Error{[}$\mathrm{m}${]}\end{tabular}}
		&\multicolumn{1}{c}{\begin{tabular}[c]{@{}c@{}}Acc@\\0.2$\mathrm{m}${[}\%{]}\end{tabular}} &\multicolumn{1}{c}{CD{[}$\mathrm{m}${]}} & \multicolumn{1}{c}{\begin{tabular}[c]{@{}c@{}}F-score\\@0.2$\mathrm{m}$\end{tabular}} \\	\cline{1-6}		
		\multicolumn{1}{c}{\begin{tabular}[c]{@{}c@{}}MapRayCasting\end{tabular}}
		&\multicolumn{1}{c}{-}
		&\multicolumn{1}{c}{0.898}  &\multicolumn{1}{c}{34.812}  &\multicolumn{1}{c}{0.452}  &\multicolumn{1}{c}{0.729} \\
		\multicolumn{1}{c}{OriginalNeRF} & \multicolumn{1}{c}{one-step}  &\multicolumn{1}{c}{2.441} &\multicolumn{1}{c}{8.290} &\multicolumn{1}{c}{1.438}  &\multicolumn{1}{c}{0.271} \\
		\multicolumn{1}{c}{\textbf{PC-NeRF}} & \multicolumn{1}{c}{two-step} &\multicolumn{1}{c}{\textbf{0.658}} &\multicolumn{1}{c}{\textbf{43.556}} &   \multicolumn{1}{c}{\textbf{0.273}} &\multicolumn{1}{c}{\textbf{0.834}}  \\   \cline{1-6}	                             
	\end{tabularx}
	}
	\label{tab:table-KITTI-sequence03}%
\end{table}

\subsection{Evaluation for 3D Reconstruction under Sensor Data Loss}\label{sec:Lesser point clouds for 3D reconstruction}

We evaluate our proposed PC-NeRF's 3D reconstruction under sensor data loss on the MaiCity 00 and KITTI 00 sequences. In real autonomous vehicle applications, some unfavorable conditions, such as hardware failures, adverse weather factors, and unstable communication in remote control tasks, may result in different degrees of sensor data loss. To simulate this challenge, we control the sensor data loss rate by setting the proportion of training point cloud data and using all the other point cloud data as the test set.

Let us first concentrate on the number of training epochs since the fewer the number of training epochs used to meet the required 3D reconstruction accuracy, the faster our proposed PC-NeRF can be deployed. As seen in Tab. \ref{tab:Lesser} and Fig. \ref{fig:fig_motivation2}, with only one epoch training, our proposed PC-NeRF can reconstruct urban and rural road environments very well until the sensor data loss rate reaches 67$\%$. However, due to too little sensor data input for model training when increasing the sensor data loss rate from 67$\%$, our proposed PC-NeRF that continues to be trained using only one epoch fails to infer using the two-step depth inference. Therefore, we increase the number of training epochs until successful inference is achieved using the two-step depth inference. At the sensor data loss rate of 90$\%$, the lowest number of epochs required for the two-step depth inference to be effective is five for scenes from the MaiCity dataset and ten for scenes from the KITTI dataset. This difference in epoch number is because the scenes in the KITTI and MaiCity datasets are rural and urban road scenes, respectively. Urban road scenes, with more straightforward environmental geometric segments, require fewer training epochs for efficient learning.

Let us then focus on the 3D reconstruction accuracy of our proposed PC-NeRF. As Tab. \ref{tab:Lesser} shows, using several training epochs that are no more or even much fewer than that of OriginalNeRF, our proposed PC-NeRF achieves much higher accuracy than OriginalNeRF and also higher accuracy than MapRayCasting. Surprisingly, in Tab. \ref{tab:Lesser}, we can find that as the sensor data loss rate increases, the 3D reconstruction accuracy of our proposed PC-NeRF does not decrease but instead increases or remains relatively stable. This accuracy does not decrease because our proposed PC-NeRF can rapidly learn an approximate scene distribution at the segment level. Moreover, if the input sensor data is too much or too little during training, more training epoch is needed to fit the model. Additionally, the 3D reconstruction accuracy is higher on the MaiCity dataset than on the KITTI dataset because urban roads have fewer environmental geometric segments than rural roads, contributing more to segment-level 3D reconstruction and, in turn, to scene-level and point-level 3D reconstruction.

In summary, our proposed PC-NeRF can effectively address the challenge of sensor data loss. Even when up to 67$\%$ of the sensor data is lost, our proposed PC-NeRF can still achieve a high novel LiDAR view synthesis accuracy (e.g., Avg. Error $<$ 0.44$\,\mathrm{m}$) and 3D reconstruction accuracy (e.g., CD $<$ 0.20$\,\mathrm{m}$) using only one training epoch, presenting a good potential for deployment and reconstruction performance even under partial sensor data loss.

\subsection{Ablation Study}\label{sec:Ablation}
To demonstrate the effectiveness of our method components, we conducted experiments on different components of our proposed PC-NeRF. 

\textbf{Effect of child NeRF free loss}: In our proposed PC-NeRF, the child NeRF free loss controlled by $\lambda_{\mathrm{cf}}$ optimizes the scene-level and segment-level environmental representation. As seen in Tab. \ref{tab:table1}, the child NeRF free loss helps improve 3D reconstruction accuracy within a specific range of variation, but too high or too low reduces the 3D reconstruction accuracy.

\textbf{Effect of balancing child NeRF free loss and child NeRF depth loss}: Based on child NeRF free loss, child NeRF depth loss is used to optimize the point-level and segment-level environmental representations further, making the balance between child NeRF free loss and child NeRF depth loss a critical point. As shown in Tab. \ref{tab:table3}, enlarging the smooth transition interval $\gamma$ between child NeRF free loss and child NeRF depth loss can improve the 3D reconstruction accuracy. At the same time, too large a smooth transition interval will decrease the 3D reconstruction accuracy. This accuracy decrease is because when the smooth transition interval is too large, child NeRF depth loss almost needs to supervise the sampling points on the entire ray and cannot effectively supervise the sampling points around the near and far bounds of child NeRF.

\textbf{Effect of two-step depth inference}: In Tab. \ref{tab:Parent-child NeRF Inference Effect}, it can be seen that two-step depth inference outperforms one-step depth inference in many cases, but inference errors also occur for OriginalNeRF. This inference error is because the validity of two-step depth inference is based on the model output weight distribution near the geometric segments' distribution in the environment, as shown in Fig. \ref{fig_infer}(b). The extensive experiments in Sec.~\ref{sec:Evaluating}, Sec.~\ref{sec:Lesser point clouds for 3D reconstruction}, and Sec.~\ref{sec:Ablation} demonstrate that our proposed PC-NeRF training method and two-step depth inference are highly compatible and consistently robust.

\begin{table*}[!t]
	\centering
	\renewcommand\arraystretch{1.4}			
	\caption{3D reconstruction under sensor data loss}		
	\begin{tabularx}{\textwidth}{>{\centering\arraybackslash}p{0.15in}>{\centering\arraybackslash}p{0.15in}|>{\centering\arraybackslash}p{0.15in}>{\centering\arraybackslash}p{0.15in}>{\centering\arraybackslash}p{0.15in}>{\centering\arraybackslash}p{0.15in}>{\centering\arraybackslash}p{0.15in}|>{\centering\arraybackslash}p{0.15in}>{\centering\arraybackslash}p{0.15in}>{\centering\arraybackslash}p{0.15in}>{\centering\arraybackslash}p{0.15in}|>{\centering\arraybackslash}p{0.15in}>{\centering\arraybackslash}p{0.15in}>{\centering\arraybackslash}p{0.15in}>{\centering\arraybackslash}p{0.15in}>{\centering\arraybackslash}X}		
		\cline{1-16}
		\midrule
		\multicolumn{1}{c}{\begin{tabular}[c]{@{}c@{}}Sensor\\data loss\\ratio[\%]\end{tabular}} & \multicolumn{1}{c|}{\begin{tabular}[c]{@{}c@{}}Train\\dataset\\ratio\end{tabular} } & \multicolumn{1}{c}{\begin{tabular}[c]{@{}c@{}}Train\\epoch\\num\end{tabular}} & \multicolumn{1}{c}{\begin{tabular}[c]{@{}c@{}}Avg.\\Error\\{[}$\mathrm{m}${]}\end{tabular}} & \multicolumn{1}{c}{\begin{tabular}[c]{@{}c@{}}Acc@\\1$\mathrm{m}${[}\%{]}\end{tabular}} & \multicolumn{1}{c}{\begin{tabular}[c]{@{}c@{}}CD\\{[}$\mathrm{m}${]}\end{tabular}} & \multicolumn{1}{c|}{\begin{tabular}[c]{@{}c@{}}F-score\\@$1\mathrm{m}$\end{tabular}} &\multicolumn{1}{c}{\begin{tabular}[c]{@{}c@{}}Avg.\\Error\\{[}$\mathrm{m}${]}\end{tabular}} & \multicolumn{1}{c}{\begin{tabular}[c]{@{}c@{}}Acc@\\1$\mathrm{m}${[}\%{]}\end{tabular}} & \multicolumn{1}{c}{\begin{tabular}[c]{@{}c@{}}CD\\{[}$\mathrm{m}${]}\end{tabular}} & \multicolumn{1}{c|}{\begin{tabular}[c]{@{}c@{}}F-score\\@$1\mathrm{m}$\end{tabular}} &\multicolumn{1}{c}{\begin{tabular}[c]{@{}c@{}}Train\\epoch\\num\end{tabular}} &\multicolumn{1}{c}{\begin{tabular}[c]{@{}c@{}}Avg.\\Error\\{[}$\mathrm{m}${]}\end{tabular}} & \multicolumn{1}{c}{\begin{tabular}[c]{@{}c@{}}Acc@\\1$\mathrm{m}${[}\%{]}\end{tabular}} & \multicolumn{1}{c}{\begin{tabular}[c]{@{}c@{}}CD\\{[}$\mathrm{m}${]}\end{tabular}} & \multicolumn{1}{c}{\begin{tabular}[c]{@{}c@{}}F-score\\@$1\mathrm{m}$\end{tabular}}	 \\
		\cline{1-16}
		\midrule		
		\multicolumn{2}{c|}{} & \multicolumn{14}{c}{\textbf{MaiCity-sequence00-0-49 (urban road)}} \\
		\cline{3-16}
		\multicolumn{2}{c|}{} & \multicolumn{5}{c|}{\textbf{PC-NeRF (two-step depth inference)}} & \multicolumn{4}{c|}{\textbf{MapRayCasting}} & \multicolumn{5}{c}{\textbf{OriginalNeRF (one-step depth inference)}} \\
		\cline{1-16}
		20    & 4/5   & \multicolumn{1}{c}{\multirow{5}[5]{*}{\centering\textbf{1}}} & 0.303 & 93.580 & 0.172 & 0.985 & 0.841 & 78.040 & 0.600 & 0.917 & \multicolumn{1}{c}{\multirow{7}[5]{*}{\centering\textbf{10}}} & 3.357 & \textbf{0.247} & 3.184 & 0.224 \\
		25  & 3/4   &       & 0.287 & 93.874 & 0.166 & 0.986 & 0.825 & 78.224 & 0.597 & 0.916 &       & 3.440 & 0.178 & 3.289 & 0.213 \\
		33.33 & 2/3   &       & 0.328 & 93.439 & 0.192 & 0.983 & 0.820 & 78.112 & 0.594 & 0.916 &       & 3.809 & 0.078 & 3.688 & 0.147 \\
		50    & 1/2   &       & 0.245 & 94.531 & 0.143 & 0.990 & 0.801 & 78.552 & 0.581 & 0.919 &       & 3.512 & 0.124 & 3.372 & 0.202 \\
		\textbf{66.67} & 1/3   &       & \textbf{0.189} & \textbf{95.448} & \textbf{0.109} & \textbf{0.994} & 0.717 & 79.360 & 0.530 & 0.929 &       & 3.715 & 0.080 & 3.633 & 0.164 \\
		\cline{3-7}    80    & 1/5   & \multicolumn{1}{c}{\multirow{2}[2]{*}{\centering\textbf{5}}} & 0.177 & 95.657 & 0.106 & 0.995 & \textbf{0.473} & \textbf{86.118} & \textbf{0.367}  & \textbf{0.972} &       & \textbf{3.356} & 0.232 & \textbf{3.174} & \textbf{0.227} \\
		90    & 1/10  &       & 0.163 & 95.853 & 0.108 & 0.997 & 0.561 & 84.773 & 0.423 & 0.956 &       & 3.406 & 0.195 & 3.218 & 0.219 \\ 
		\cline{1-16}
		\midrule
	    \multicolumn{2}{c|}{} & \multicolumn{14}{c}{\textbf{KITTI-sequence00-1151-1200 (rural road)}} \\
   		\cline{3-16}
	    \multicolumn{2}{c|}{} & \multicolumn{5}{c|}{\textbf{PC-NeRF (two-step depth inference)}} & \multicolumn{4}{c|}{\textbf{MapRayCasting}} & \multicolumn{5}{c}{\textbf{OriginalNeRF (one-step depth inference)}} \\
   		\cline{1-16}
	    20    & 4/5   & \multicolumn{1}{c}{\multirow{5}[0]{*}{\textbf{1}}} & 0.488 & 92.131 & 0.224 & 0.993 & 0.871 & 77.823 & 0.466 & 0.949 & \multicolumn{1}{c}{\multirow{7}[0]{*}{\textbf{10}}} & 5.251 & 1.515 & 3.289 & 0.321 \\
	    25  & 3/4   &       & 0.443 & 93.391 & 0.206 & \textbf{0.995} & 0.865 & 78.146 & 0.463  & 0.949 &       & 5.098 & 1.743 & 3.026 & 0.346 \\
	    33.33 & 2/3   &       & 0.447 & 93.296 & 0.206 & \textbf{0.995} & 0.858 & 78.449 & 0.459  & 0.950 &       & 5.361 & 1.392 & 3.166 & 0.300 \\
	    50    & 1/2   &       & 0.461 & 92.860 & 0.213 & 0.994 & 0.851 & 78.209 & 0.459  & 0.950 &       & 4.968 & 2.182 & 3.107 & 0.361 \\
	    \textbf{66.67} & 1/3   &  & \textbf{0.439} & \textbf{93.558} & \textbf{0.197} & \textbf{0.995} & 0.855 & 78.297 & 0.459  & 0.949 &       & 5.561 & 1.273 & 3.687 & 0.264 \\
	    \cline{3-7}
	    80    & 1/5   &  \multicolumn{1}{c}{\centering\textbf{5}}    & 0.423 & 93.661 & 0.197 & 0.995 & 0.813 & 79.263 & 0.437  & 0.955 &       & \textbf{2.487} & \textbf{17.934} & \textbf{1.301} & \textbf{0.858} \\
	    \cline{3-7}
	    90    & 1/10  & \multicolumn{1}{c}{\centering\textbf{10}} & 0.412 & 93.665 & 0.197 & 0.995 & \textbf{0.780} & \textbf{80.210} & \textbf{0.417}  & \textbf{0.960} &       & 5.195 & 2.058 & 3.015 & 0.335 \\
		\cline{1-16}
		\midrule	    		
	\end{tabularx}%
	\label{tab:Lesser}%
\end{table*}%

\begin{table}[!t]
	\centering
	\renewcommand\arraystretch{1.4}		
	\caption{Effect of child NeRF free loss}		
	\begin{tabular}{c|cccc}
		\cline{1-5}
		\midrule		
		\multicolumn{1}{c|}{$\lambda_{\mathrm{cf}}$}     & \multicolumn{1}{c}{Avg. Error[$\mathrm{m}$]} & \multicolumn{1}{c}{Acc@$0.2\mathrm{m}$[\%]} & \multicolumn{1}{c}{CD[$\mathrm{m}$]} & \multicolumn{1}{c}{F-score@$0.2\mathrm{m}$} \\
		\cline{1-5}
		\midrule		
		0     & 0.5110  & 65.2321  & 0.2201  & 0.8904  \\
		1     & 0.5072  & 66.1292  & 0.2165  & 0.8939  \\
		10    & 0.4930  & 68.3496  & 0.2099  & 0.8995  \\
		100   & 0.4743  & \textbf{70.0247}  & \textbf{0.2067}  & \textbf{0.9000}  \\
		1000  & 0.4723  & 69.5482  & 0.2092  & 0.8987  \\
		$10^{4}$ & 0.4646  & 69.9601  & 0.2087  & 0.8984  \\
		$10^{5}$ & \textbf{0.4633}  & 67.1306  & 0.2182  & 0.8936  \\
		$10^{6}$ & 0.4647  & 67.1092  & 0.2185  & 0.8935  \\
		$10^{7}$ & 0.4673  & 67.0546  & 0.2192  & 0.8931  \\
		$10^{8}$ & 0.4663  & 67.0783  & 0.2189  & 0.8933  \\
		$10^{9}$ & 0.4655  & 67.0933  & 0.2187  & 0.8934  \\
		\cline{1-5}		
		\multicolumn{5}{l}{Note: $\lambda_{\mathrm{cd}} = \lambda_{\mathrm{in}} = \gamma = 0$} \\
		\cline{1-5}
		\midrule		
	\end{tabular}%
	\label{tab:table1}%
\end{table}%

\begin{table}[!t]
	\centering
	\renewcommand\arraystretch{1.4}			
	\caption{Effect of balancing child NeRF free loss and child NeRF depth loss}		
	\begin{tabular}{c|cccc}	
		\cline{1-5}
		\midrule			
		\multicolumn{1}{c|}{$\gamma[\mathrm{m}]$} & \multicolumn{1}{c}{Avg. Error[$\mathrm{m}$]} & \multicolumn{1}{c}{Acc@$0.2\mathrm{m}$[\%]} & \multicolumn{1}{c}{CD[$\mathrm{m}$]} & \multicolumn{1}{c}{F-score@$0.2\mathrm{m}$} \\
		\cline{1-5}
		\midrule			
		\multicolumn{1}{c|}{0.25} & \multicolumn{1}{c}{0.5113 } & \multicolumn{1}{c}{66.4901 } & \multicolumn{1}{c}{0.2283 } & \multicolumn{1}{c}{0.8870 } \\
		\multicolumn{1}{c|}{0.5} & \multicolumn{1}{c}{0.5113 } & \multicolumn{1}{c}{66.4303 } & \multicolumn{1}{c}{0.2282 } & \multicolumn{1}{c}{0.8873 } \\
		\multicolumn{1}{c|}{1} & \multicolumn{1}{c}{0.5130 } & \multicolumn{1}{c}{65.9698 } & \multicolumn{1}{c}{0.2286 } & \multicolumn{1}{c}{0.8874 } \\
		\multicolumn{1}{c|}{1.5} & \multicolumn{1}{c}{0.5120 } & \multicolumn{1}{c}{65.8837 } & \multicolumn{1}{c}{0.2290 } & \multicolumn{1}{c}{0.8878 } \\
		\multicolumn{1}{c|}{2} & \multicolumn{1}{c}{0.4884 } & \multicolumn{1}{c}{66.6541 } & \multicolumn{1}{c}{0.2239 } & \multicolumn{1}{c}{0.8908 } \\
		\multicolumn{1}{c|}{2.5} & \multicolumn{1}{c}{\textbf{0.4446} } & \multicolumn{1}{c}{68.2699 } & \multicolumn{1}{c}{0.2097 } & \multicolumn{1}{c}{0.8979 } \\
		\multicolumn{1}{c|}{3} & \multicolumn{1}{c}{0.4474 } & \multicolumn{1}{c}{68.6973 } & \multicolumn{1}{c}{0.2073 } & \multicolumn{1}{c}{0.8988 } \\
		\multicolumn{1}{c|}{4} & \multicolumn{1}{c}{0.4492 } & \multicolumn{1}{c}{69.0594 } & \multicolumn{1}{c}{\textbf{0.2063} } & \multicolumn{1}{c}{0.8992 } \\
		\multicolumn{1}{c|}{5} & \multicolumn{1}{c}{0.4517 } & \multicolumn{1}{c}{69.3440 } & \multicolumn{1}{c}{\textbf{0.2063} } & \multicolumn{1}{c}{\textbf{0.8994} } \\
		\multicolumn{1}{c|}{7.5} & \multicolumn{1}{c}{0.4662 } & \multicolumn{1}{c}{\textbf{70.5668} } & \multicolumn{1}{c}{0.2068 } & \multicolumn{1}{c}{0.8993 } \\
		\multicolumn{1}{c|}{10} & \multicolumn{1}{c}{0.4683 } & \multicolumn{1}{c}{70.2688 } & \multicolumn{1}{c}{0.2086 } & \multicolumn{1}{c}{0.8987 } \\
		\multicolumn{1}{c|}{20} & \multicolumn{1}{c}{0.4704 } & \multicolumn{1}{c}{69.7122 } & \multicolumn{1}{c}{0.2081 } & \multicolumn{1}{c}{0.8993 } \\
		\cline{1-5}		
		\multicolumn{5}{l}{Note: $\lambda_{\mathrm{cf}} = 10^{6}, \lambda_{\mathrm{cd}} = 10^{5}, \lambda_{\mathrm{in}} = 0.1$} \\
		\cline{1-5}
		\midrule			
	\end{tabular}%
	\label{tab:table3}%
\end{table}%

\section{Conclusion}
This paper proposes a large-scale, high-precision scene reconstruction framework called parent-child neural radiance fields (PC-NeRF), effectively improving reconstruction performance under partial sensor data loss in outdoor environments. The two modules of PC-NeRF, namely, the parent NeRF and the child NeRF, provide volumetric representations for the entire autonomous vehicle traversal environments and individual scene geometric segments. Using LiDAR rays intersecting parent NeRF and child NeRF, we design three losses, including the parent NeRF depth loss, child NeRF free loss, and child NeRF depth loss, to jointly optimize the scene-level, segment-level, and point-level scene representation. Our proposed PC-NeRF is validated with extensive experiments to achieve high-precision 3D reconstruction in large-scale scenes. Moreover, PC-NeRF has strong deployment potential, with considerable scene representation accuracy that can be achieved by training only one epoch in most test scenarios. Most importantly, PC-NeRF effectively tackles partial sensor data loss problems in real vehicle applications. Our future work will further explore the potential of PC-NeRF in object detection and localization for autonomous driving.

\bibliographystyle{IEEEtran}      
\bibliography{reference}                        

\end{document}